\newcommand{\cost}{\mathrm{cost}}

\documentclass[12pt]{article}
\usepackage{amsmath,amssymb}
\usepackage{graphicx}
\usepackage{color}
\newcommand{\changed}[1]{{#1}}

\usepackage{dcolumn}
\usepackage{bm}
\usepackage{float}
\usepackage{hyperref} 
\usepackage{apacite}
\usepackage{adjustbox}
\usepackage{makecell}

\hypersetup{ colorlinks = true, citecolor = black, urlcolor = blue}
\newcommand{\qed}{\hfill \ensuremath{\Box}}

\definecolor{background-color}{gray}{0.98}
\newcommand{\br}[1]{\left\{#1\right\}}                            

\newcommand{\REAL}{\ensuremath{\mathbb{R}}}
\providecommand{\norm}[1]{\left\lVert#1\right\rVert}
\newcommand{\eps}{\varepsilon}
\renewcommand{\epsilon}{\varepsilon}

\date{}
\renewcommand{\paragraph}[1]{\medskip\noindent\textbf{{#1}}}
\title{Introduction to Core-sets: an Updated Survey}
\author{Dan Feldman \thanks{Computer Science Department, University of Haifa. Figures by Ibrahim Jubran.}}
\begin{document}
\maketitle
\begin{abstract}
In optimization or machine learning problems we are given a set of items, usually points in some metric space, and the goal is to minimize or maximize an objective function over some space of candidate solutions. For example, in clustering problems, the input is a set of points in some metric space, and a common goal is to compute a set of centers in some other space (points, lines) that will minimize the sum of distances to these points. In database queries, we may need to compute such a some for a specific query set of $k$ centers.


However, traditional algorithms cannot handle modern systems that require parallel real-time computations of infinite distributed streams from sensors such as GPS, audio or video that arrive to a cloud, or networks of weaker devices such as smartphones or robots.

Core-set is a ``small data" summarization of the input ``big data", where every possible query has approximately the same answer on both data sets. Generic techniques enable efficient coreset \changed{maintenance} of streaming, distributed and dynamic data. Traditional algorithms can then be applied on these coresets to maintain the approximated optimal solutions.

The challenge is to design coresets with provable tradeoff between their size and approximation error.
This survey summarizes such constructions in a retrospective way, that aims to unified and simplify the state-of-the-art.
\end{abstract}

\section{Motivation}
Every day 2.5 exabytes of data ($2.5\times 10^{18}$)~\cite{IBM} are generated by cheap and numerous information-sensing mobile devices, remote sensing, software logs, cameras, microphones, RFID readers and wireless sensor networks~\cite{segaran2009beautiful,hellerstein2008parallel,funke2007bounded}. These require clustering algorithms that, unlike traditional algorithms, (a) learn unbounded streaming data that cannot fit into main memory, (b) run in parallel on distributed data among thousands of machines, (c) use low communication between the machines (d) apply real-time computations on the device,
(e) handle privacy and security issues.

A common approach is to re-invent computer science for handling these new computational models, and develop new algorithms ``from scratch" independently of existing solutions.

\newcommand{\far}{\mathrm{far}}
\begin{figure}[!ht]
\centering
\includegraphics[scale=0.2,width=0.3\textwidth]{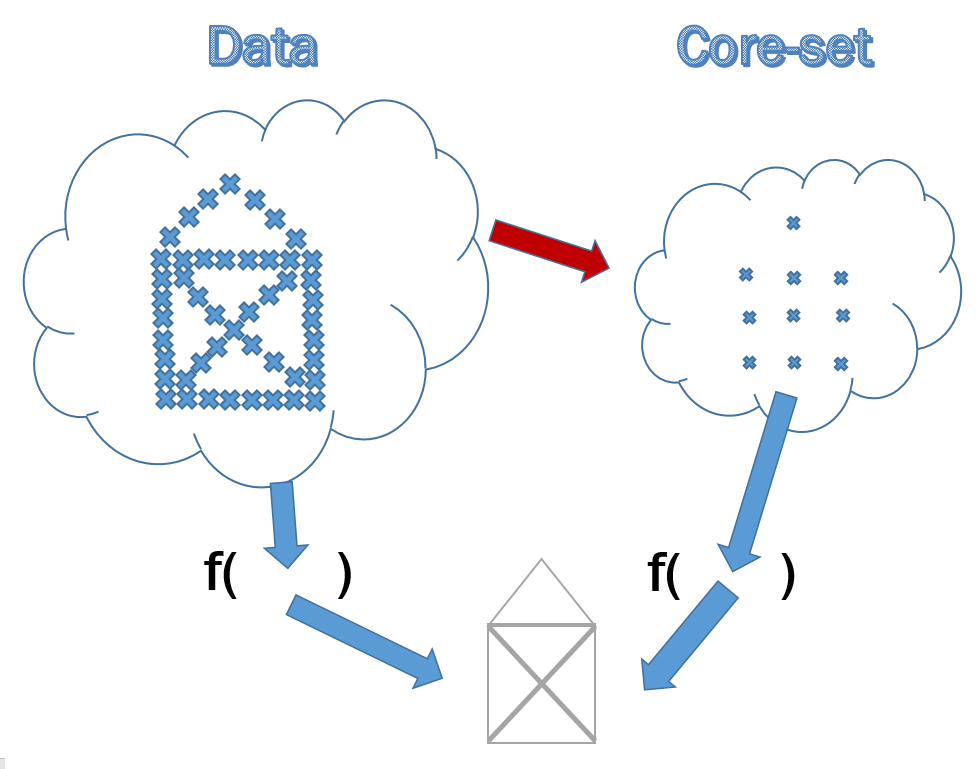}
\includegraphics[scale=0.2,width=0.3\textwidth]{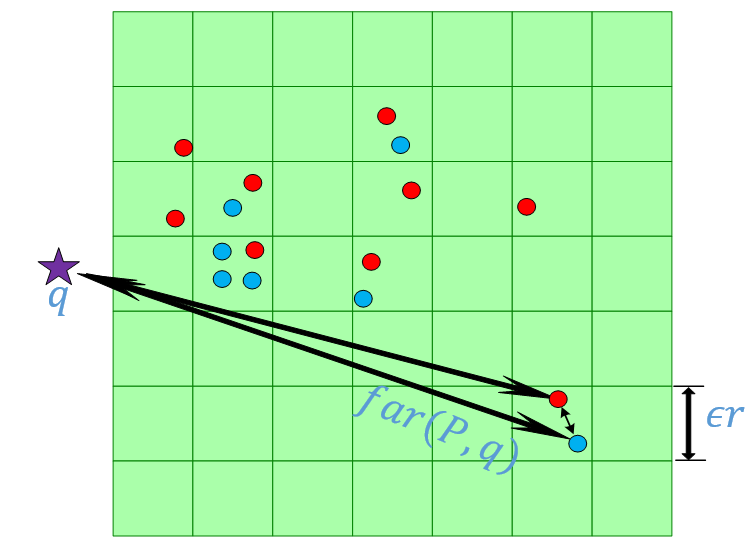}
\includegraphics[scale=0.2,width=0.3\textwidth]{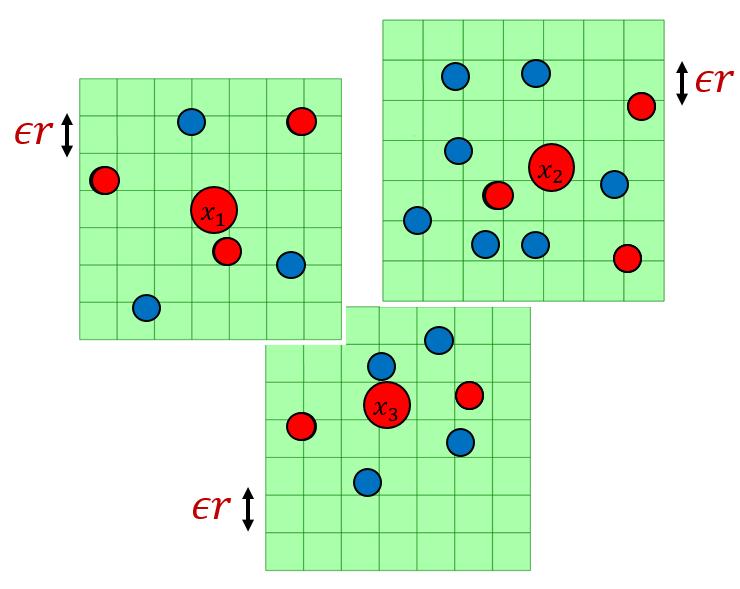}
\caption{\scriptsize\label{figa} \textbf{(left)} A function (algorithm) $f$ gets a points set and outputs its clustering into linear segments. Applying $f$ on the coreset of the data would ideally yield the same result faster using less resources, without changing $f$. \textbf{(middle)} An $\eps$-coreset (in red) for $1$-center queries $\far(P,q)=\max_{p\in P}\norm{p-q}_2$ of a set $P$ of (blue) points and a query $q$, both on the plane, where $r=\min_{q'\in \REAL^2}\far(P,q')$ is the minimum cost.
\textbf{(right)} In $k$-center clustering, the query is replaced by a set $Q$ of $k$ centers, $\far(P,Q)=\max_{p\in P}\min_{q\in Q}\norm{p-q}_2$ and a grid is constructed around each of the optimal centers.}
\end{figure}

Data reduction/summarization advocates a different approach: given a large data set, reduce the data so that it takes up significantly less space in memory, but \emph{provably} approximates the large data in a problem dependent sense. Running existing (possibly off-line, non-parallel or inefficient) algorithms on the reduced (small) data would then produce a result that is provably close to the solution obtained from running on the complete (big) data. Such a compressed data set is sometimes called a core-set (or, coreset); see Fig.~\ref{figa}(left).

\paragraph{Zen coresets }might be a better name for the coresets due to their flexible definition, based on the problem and related community: when we can't find a coreset for a problem, we change the definition of a coreset. This is part of the reason why curious researchers and engineers find it hard to get into the coreset ``cult".

\paragraph{The coreset paradigm }shift is another source of confusion between both readers, users, coreset developers, and especially reviewers: why your coreset construction algorithm assumes that it gets the optimal solution as input, if this is the main motivation for constructing it in the first place? Why you claim $O(n)$ construction time when the main theorem states $O(n^5)$? Where are the streaming and parallel coreset constructions (there are only off-line constructions)? Where is the analysis for the communication between the machines? What is wrong with uniform sampling as a coreset? Why do we need coresets, if Gradient Descent provides a sparse solution in linear time? What is the generalization error? Why there are experiments only on training (not testing) data?

The goal of this survey is to answer such questions by introducing the coreset paradigm together with the recent frameworks, as well as explaining their relations to other techniques.

\paragraph{Structure of this paper.} The paper begins with high-level discussions and general definitions, continues to frameworks, and then to specific problems and solutions. Section~\ref{sec:what} presents several inconsistent definitions of coresets that were suggested over the years, by focusing on their different properties, and provide a definition that captures most of them. Section~\ref{sec:why} suggests example scenarios where coresets can be used or combined with existing solutions. Section~\ref{sec:prob} defines the type of problems that coreset may be applied on with some examples, while Section~\ref{sec:types} defines specific types of coresets to solve such problems.

While there are dozens or hundreds of coreset constructions, Section~\ref{sec:generic} aims to present a framework called sup-sampling that can be used to solve many of them, by unifying existing frameworks. It usually reduces the coreset construction to the computation of the $\sup$ or $\max$ of corresponding functions, as well as their corresponding dimension (that is related to the notion of VC-dimension). To this end, Section~\ref{sec:bound} survey techniques to bound $\sup$ and dimension for common family of problems.

Section~\ref{future} suggests future research and open problems. 


\newcommand{\ariel}[1]{#1}
\section{What is a Coreset?\label{sec:what}}
The term coreset was coined in~\cite{agarwal2005geometric} and used for computing the smallest $k$ balls that cover a set of input points, and then similar covering problems where coresets are called certificates (e.g. in~\cite{agarwal2002approximation}); see survey in~\cite{agarwal2005geometric}. Today there are many inconsistent coreset definitions and data reduction techniques adding to the motivation for writing this survey. We focus on coresets construction with provable guarantees for the trade-offs between size of coreset and approximation error. Other parameters such as construction or update time are usually derived from this trade-off; see Section~\ref{sec:generic}.
\newcommand{\dist}{f}

\paragraph{Query space }is a tuple $(P,w,\mathcal{X},f)$, where $P$ is a (usually finite and of size $|P|=n$) input set that is called \emph{points}, $\mathcal{X}$ is a (usually infinite) set of \emph{queries} (models, shapes, classifiers, hypotheses), $w:P\to \REAL$ is a (usually positive) \emph{weight} function that assigns importance to each point, and $f:P\times \mathcal{X}\to\REAL$ is a (usually non-negative) \emph{pseudo-distance} function, or distance for short $f(p,x)$ from $p$ to $x$; see~\cite{FL11,braverman2016new}. More generally it can be considered as a loss function. For max optimization problem it may used as score (where higher is better).

\paragraph{Coreset }in this review, is a small data structure $C$ that allows us to approximate the sum of weighted distances $\sum_{p\in P}w(p)f(p,x)$, its maximum distance $\max_{p\in P}w(p)f(p,x)$ which is related to \emph{covering problems}\ariel{, or in} general, any function $\cost:\REAL^n\to \REAL$ that maps the $n=|P|$ distances $(f(p,x))_{p\in P}$ to a (usually non-negative) total cost.
Exact definitions for ``small", ``data structure", and ``approximates" are changed from paper to paper and we suggest \ariel{a few} of them in Section~\ref{sec:types}.

\paragraph{Example: }$P\subseteq\REAL^d$, $w\equiv 1$, $\mathcal{X}=\br{X\subseteq \REAL^d\mid |X|=k}$, and $f(p,X)=\min_{x\in X}\norm{p-x}_2$ corresponds to $k$-means queries that aims to compute an \emph{optimal query} set $X^*$ of $|X^*|=k$ centers that minimizes the sum of squared distances $\sum_{p\in P}f^2(p,X)=\norm{(f(p,X))_{p\in P}}^2_2$ over $X\in\mathcal{X}$. 
More generally, $\REAL^d$ can be replaced by any set, and $\norm{p-x}_2$ by any distance $\dist(p,x)$.In the $k$-median problem the cost is the non-squared distances $\sum_{p\in P}f(p,X)=\norm{(f(p,X))_{p\in P}}_1$.

\paragraph{Composable coresets}~\cite{bent,HM04,indyk2014composable} are important types of coresets that make them especially relevant for handling
big or real-time data as explained in the next section. It means that a union of a pair of coresets $C_1\cup C_2$ is a coreset for the underlying input $P_1\cup P_2$, and that we can re-compute coreset $C_3$ for this coreset $C_1\cup C_2$ recursively. In particular, strong and weak coresets as defined in Subsection~\ref{sec:query} are composable. Using merge-and-reduce trees~\cite{bent} we can apply the coreset construction only on small subsets of the input independently (i.e., ``embarrassingly parallel"~\cite{foster1995designing}), merge them, and then reduce recursively.  This implies that an efficient off-line coreset construction can be applied only \ariel{to} small subsets, to obtain linear coreset construction time via sub-linear memory; see Section~\ref{sec:cons}.

\section{Why coresets?\label{sec:why}}
In this section we give only a few of the many advantages of using coresets, or at least, composable coresets.

\paragraph{Answering queries, }\ariel{such} as SQL queries, to save either time, memory or communication via the compressed small coreset $C$, including on the computation models below.

\paragraph{Optimization }is the most common motivation for constructing coresets, where the goal is to compute an optimal query that minimizes the cost. Solving the optimization problem or its approximation on the small coreset yields an approximate solution of the original (big) dataset, sometimes after \ariel{suitable} post-processing. For example, the $k$-means problem is NP-hard when $k$ is part of the input~\cite{mahajan2009planar}. However, composable coresets of \ariel{near-linear size} in $(k/\eps)$ can be used to produce $1\pm\eps$ multiplicative factor approximation in $O(ndk)$ time, even for streaming distributed data in parallel~\cite{braverman2016new,FSS13}.

\paragraph{Boosting existing heuristics,} i.e. algorithms with no provable guarantees
is \ariel{possible} by running them on the small coreset. For example, \ariel{we can} run many more iterations or initial seeds \ariel{on the coreset} in the same time it takes for a single run on the original (big) data. In this sense we ``improve the state-of-the-art using the state-of-the-art", and coreset is used as a bridge between theory and practical systems; see examples in~\cite{paul2014visual,feldman2013idiary,epstein2018quadcopter}.

\paragraph{``Magically" turn existing off-line algorithms to }(i) streaming algorithms, i.e, that use small memory and one pass over a possibly infinite input stream~\cite{bent,HM04}, (ii) parallel algorithms that use multiple threads (as in GPU), or more
generally distributed data (network/cloud/swarm of robots or smartphones) via low or no
communication between the machines~\cite{indyk2014composable}, also simultaneously for unbounded stream of data~\cite{FT15} as in (i), and (iii) dynamic (insertions \& deletions of points) in small time, but using linear space~\cite{AHV04}.

This is done by maintaining a single coreset via these computational models during the night (or every few seconds). In the morning (or every few seconds) we apply the possibly inefficient existing algorithm in an off-line, non-parallel way from scratch on this coreset that represents all the data.

\paragraph{Constrained optimization }can be computed on a coreset that was constructed independently of these constraints. E.g. coreset for $k$-means queries can be used to compute the optimal query whose centers must be subset of the input, are close to some input points, or cannot be placed in forbidden areas, by solving the constrained problem on the coreset.

\paragraph{Model selection and features reduction }can be computed and evaluated on the coreset in a similar way.
A particular useful property is that a coreset for, say, $k$ centers is a coreset for $k' \leq k$ centers by definition: put $k$ centers on top of $k'$ centers. This allows us to handle regularization terms or costs that are common in machine learning such as to compute $X$ that minimizes $\cost_f(P,X)+|X|$ over every set $X$ of $|X|\leq k$ centers~\cite{agakmean,bachem2015coresets}. Same property occurs for other model parameters such as $j$-subspaces;~\cite{FSS13}.

\changed{\paragraph{Disadvantages. }There are few disadvantage of the coresets paradigm as follows:
\begin{enumerate}
\item Coresets may simply do not exist for the problem at hand. This is usually due to too high supremums/sensitivities or related VC-dimension of the query space as explained in the next sections. For example, we should not expect a coreset of nodes for the problem of computing the shortest path between a pair of nodes in a graph, or a coreset for a nearest neighbour, at least for the simplest coreset definitions. In these cases, we usually suggest a relaxed definition of the problem or coreset. This may be the reason that most of the known coresets are related to data science or machine/deep learning problems, where we try to represent large input data with a relatively small model.
\item Coreset schemes may be hard to design. A coreset construction scheme for a new problem may be very similar to such an existing construction. However, it took a few years to come up with at least some of the coresets constructions that are referenced in this paper. In this sense, coresets are similar to the research of optimization or approximation algorithms, when we do not expect a single technique that will solve all existing problems, and each problem might require its own approach. Nevertheless, in the recent years the author of this survey and many of its colleagues aim to suggest more general techniques and not specific coresets, which is analogous to recipes in optimization such as convex/linear/quadratic programming.
\item Even if we have a coreset for a problem, it might be that the additional approximation error compared to running on the original data is too large.
\item The additional coreset construction time might be too long.
\end{enumerate}

The third disadvantage is relevant when we have an algorithm that computes the global optimal solution.
The fourth disadvantage is relevant when the existing algorithm is already very fast.
For example, linear regression might be too easy problem to solve via coresets. Unlike most of the modern machine/deep learning or computer vision problems which are hard.
Nevertheless, even in easy cases such as linear regression, coresets can be useful for parameter tuning, streaming/distributed data, and other issues that are discussed in Section~\ref{sec:why}. For example, practical and accurate coresets ($\eps=0$) for linear regression and low-rank approximation can be found in~\cite{alaa}.}

\section{Coreset types\label{sec:types}}
Example coresets and their properties for $k$-means queries can be found in Table~\ref{table:k}; see details in~\cite{braverman2016new}.
Some (although not many) coresets can be used to evaluate \emph{exactly} the desired cost function. Some examples are shown in Table~\ref{table:acc}; see more details in~\cite{nasser2015low}. The last coreset in Table~\ref{table:acc} can also be used ``as is" to compute constrained regression such as ridge, LASSO, elastic net and more that can be found e.g. in~\cite{tutz2007boosting}. This is since coresets can also be used for constrained optimization as explained in Section~\ref{sec:why}. In general, coresets are different for example by the following properties.

\paragraph{Approximation error} for the majority of coresets aims to have a multiplicative $(1+\eps)$-multiplicative approximation for the desired cost of each query $X\in\mathcal{X}$, or at least the optimal query. That is, an additive error of $\eps\cost_f(P,X)$. Grid coresets (See Section~\ref{sec:cons}) usually introduce smaller error of $\eps\cost_f(P,X^*)$, where $X^*$ is the optimal query. Weaker but sometimes smaller coresets use larger additive error that may be multiplicative under some assumption on the input (e.g. well clustered~\cite{dasgupta2000two,ostrovsky2006effectiveness}, scaled data~\cite{H04,edwards2005no}), by adding more parameters to the coreset's size that takes these parameters into account~\cite{feldman2012effective,tolochinsky2018coresets}, or lower bound on the cost for the allowed queries.

\paragraph{The size }of a coreset is usually by order of magnitude smaller that the size $n$ of the input, about near-logarithmic or even independent on $n$. Rarely the reduction is on the dimension; see Low-Dimensional coresets in Sub-section~\ref{five}.  The dependency on the approximation error $\eps\in(0,1)$ is usually polynomial in $1/\eps$. Probabilistic constructions failed with a given probability $\delta\in(0,1)$ and the coreset size usually depends poly-logarithmically on $\log(1/\delta)$ (by Hoeffding's inequality~\cite{hoeffding1963probability}), and rarely linearly on $1/\delta$ (by Markov's inequality~\cite{IKI94}). First coresets were deterministic of size exponential in their (VC-)dimension $d$~\cite{agarwal2005geometric,FFS06}, random constructions whose output is polynomial in $d$ replaced them~\cite{C09}, and independency of $d$ is sometimes possible using either weak coresets~\cite{FT15} or squared Euclidean distances~\cite{FSS13,BF16,FeldmanOR17}.

\subsection{Data types\label{five}}
Any small coreset $C$ may be used to answer queries efficiently. However, if the data type of the coreset is different from the original data type of $P$, we cannot use the same loss function $f$ on $C$. In particular, we cannot directly run existing solvers for computing the optimal query on these coresets. More complicated and different data types may help to reduce the size of the coreset, but require the design of new (possibly non-efficient) optimization algorithms.

\paragraph{Weighted subset of input } is a coreset $C\subseteq P$, where each point in $C$ may be assigned a multiplicative scalar weight. The simplest case is where all the weights equal $1$, and $C$ is just a subset, which makes sense for covering problems~\cite{AHV04,AHP01}. For approximating average of distances by few points, we can still avoid weights, e.g. by uniform sample/distribution~\cite{LLS01}. For sum of weights, the weight is positive and intuitively tells us how many input points each coreset point represents~\cite{HM04,FFS06}. Most of the solvers support weighted input or can be easily changed to support it. However, negative weights or weights $u(p,x)$ that depends also on the query might imply non-convex optimization problem or other complications~\cite{FL11,feldman2010coresets,feldman2012single}; see Fig.~\ref{figa}(right).

Advantages of weighted subsets are (i) preserved sparsity of the input, (ii) interpretability, (iii) coreset may be used (heuristically) for other problems, (iv) less numerical issues that occur when non-exact linear combination of points are used.

\paragraph{Weighted subset of input space}, where we assume that the input $P$ is from a ground set or metric space, and the coreset is from the same ground set. E.g. for $k$-means, the coreset may be a subset of $\REAL^d$ but not a subset of $P$; e.g.~\cite{HK05,rosman2014coresets}. This is related to the notion of weak $\eps$-net in computational geometry~\cite{matouaek2003new} (nothing to do with weak coresets in the next section).

\paragraph{Sketch matrices} imply that each point in the coreset is a linear combination
of the input points. The input is represented as a matrix $P\in\REAL^{n\times d}$ where every point
corresponds to a row, and the coreset construction is a ``fat" matrix ($m \ll n$)
$S\in\REAL^{m\times d}$ called \emph{sketch matrix}. The coreset (called sketch) is then $C=SP\in\REAL^{m\times d}$. Sketch matrices are generalizations of previous coresets, where $S$ is either a fat or sparse diagonal $n\times n$ matrix of the original row weights. As coresets, the term sketch is not consistent along papers. Sketch matrices may support the stronger turn-style model that allows deletion and changing single entries and not only insertion of records, in sub-linear space and without using trees. See surveys in~\cite{Phillips16,CW09}.

\paragraph{Low-dimensional coresets} are coresets that instead of having small number of points, are contained in a low dimensional space in some sense. The classic examples are low-rank approximations (SVD/PCA)~\cite{Mahoney,FSS13,CEMMP15}, JL-lemma (random projections)~\cite{boutsidis2015randomized} and well conditioned matrices~\cite{DasguptaDHKM08}. Sometimes such coresets are the first step before reducing the number of input points; see Section~\ref{sec:bound} for projections on the optimal query.

\paragraph{Generic data structures} are usually the result of combining few types of the above coresets. These are
the most harder to handle for optimization, but may be unavoidable for obtaining small coreset for some problems~\cite{feldman2012single,FL11,rosman2014coresets}.

\subsection{Query sets\label{sec:query}}
In Section~\ref{sec:what} we defined coreset with respect to a set of queries that it approximates. The
first two following types of coresets are much stronger than the third, as they are also composable.

\paragraph{Strong coreset } approximates \emph{any }given query in the given query set. They are thus composable and it is usually easy to extract from them an approximation to the optimal query of the original input $P$. See examples in~\cite{agarwal2005geometric,Phillips16,FL11}.

\paragraph{Weak coreset} is a coreset $C$ that does not approximate all the queries. Instead, it approximates a subset of queries $\mathcal{X}(C)$. If this function $\mathcal{X}(C)$ is monotonic, i.e., $C'\subseteq C$ implies $\mathcal{X}(C)\subseteq \mathcal{X}(C')$, then the techniques for constructing $\eps$-sample (see Section~\ref{sec:generic}) can be applied~\cite{FL11}. Due to the smaller query space, the sample might be smaller by order of magnitude. This is since it depends on the generalized VC-dimension of the function $\mathcal{X}$ and not the VC-dimension of the set $\mathcal{X}(P)$; e.g.~\cite{FL11,FMS07,FT15}.


\emph{Example: }The VC-dimension for the range space of $k$ balls in $\REAL^d$ is $O(dk\log k)$;\cite{FSS13}. However, a $(1+\eps)$-approximation to the optimal query (center) for the $k$-means/median/center problem is spanned by a convex combination of $k/\eps^{O(1)}$ input points in $P$~\cite{SV07,FeldmanOR17}. By defining the function $\mathcal{X}$ that maps a coreset $C$ to the union $\mathcal{X}(C)$ of all these centers, we obtain a generalized VC-dimension of $1/\eps^{O(1)}$, i.e.,  independent of $d$ for the query spaces of these problems. Generalizations hold for projective clustering $(j>0)$, and suggestion on how to compute the optimal query of the restricted set $\mathcal{X}(C)$ can be found in~\cite{FL11,braverman2016new,FMS07,FT15}.


\paragraph{Sparse solution }(query) that approximates only the optimal query can be computed using convex optimization techniques such as Frank-Wolfe algorithm~\cite{BC03,CL10}. An approximation solution, sparse or not, for a subset of the input points at hand, usually says nothing about the optimal solution of the complete data, or even after the insertion of a single new point. 
Hence, sparse solutions are not composable and thus not support all the computation models from Section~\ref{sec:why}. Sometimes it is unavoidable: such coreset for $1$-center has size $O(1/\eps)$ for $\eps\in(0,1)$, while composable coreset or any coreset that handles streaming data must be of size exponential in $d$ as was proven in~\cite{agarwal2005geometric}. Sometimes the sparse solution is the weight vector of the (strong) coreset itself~\cite{feldmanmik}.

\subsection{Construction Types} \label{sec:cons}
\paragraph{Uniform sample } from the input is probably the most common ``coreset". This is also a natural competitor of every other coreset. Unlike other constructions, uniform sampling takes time sub-linear time in the
input, which also explains why it misses small but important input points or far clusters, and does not provide $(1\pm\eps)$-multiplicative error as other coresets. Nevertheless, most of the coresets use uniform sampling after proper scaling by sup-weights which enables smaller coresets using the notion of $\eps$-sample; see Section~\ref{sec:generic} and~\cite{Cohen15,FL11,LLS01}.
%

\paragraph{Importance sampling} aims to reduce the additive error of $\eps n \cost(P,x)$ to $\eps \cost(P,x)$, i.e. $(1 \pm \eps)$-multiplicative error by replacing uniform sampling with non-uniform sample of the same size over the input space.
The main technique is to re-weight each point by its sensitivity, or sup-weight respectively, as explained in Section~\ref{sec:generic}. Then compute a ``uniform" sample from the weighted set, where each input point is replaced by duplicated points according to its new weight, e.g.~\cite{braverman2016new,LS10}.

\paragraph{Grids }are based on \changed{discretization} of the input space to small clusters, and then taking a
representative from each cluster, weighted by the number of input points in its cell. These
are the first coresets and were first used for covering problems~\cite{AP03,agarwal2005geometric}.
The additive error $\eps cost(P,X^*)\leq \eps cost(P,X)$ is usually smaller than modern coresets.
However, the time and space is exponential in $d$ due to the large number of cells, which is also the reason that deterministic constructions (that takes time exponential in $d$) are used. E.g.~\cite{HM04,AP03}; see Fig.~\ref{figa}(middle/right).

\paragraph{Greedy constructions }are used for problems with specific properties to obtain
smaller coresets, e.g. based on convex optimization. Here, in each iteration we adaptively
pick the next best point to the coreset deterministically~\cite{feldmanmik,FeldmanOR17} or based on importance sampling~\cite{ADK09}.
Such deterministic constructions may obtain coresets of size that cannot be obtained via random constructions; see Fig.~\ref{figc}(right), Section~\ref{sec:lb} and~\cite{BF16,batson2014twice}

Such deterministic or adaptive constructions may be smaller by order of magnitudes compared to other constructions. For example, the query set of all the possible convex shapes (sets) has corresponding unbounded VC-dimension, but still has
an $\eps$-sample of finite size~\cite{Chazelle}; see also Lower Bounds in Section~\ref{sec:lb}.
%

\begin{table}[ht]
\begin{adjustbox}{width=1\textwidth}
\small
\begin{tabular}{ | c | c | c | c | c | c | c | c |}
\hline
Name & Input & \makecell{Query} & \makecell{Cost\\function} & \makecell{Coreset\\properties} & \makecell{Coreset\\weights} & \makecell{Const.\\time} & \makecell{Query\\time}\\
\hline
\makecell{$1$-center} & $P\subseteq \ell \in \REAL^d$ & $x\in\REAL^d$ & $\max_{p\in P}|p-x|$ & $\makecell{C\subseteq P\\ |C|=2}$ & Unweighted & $O(n)$ & $O(d)$\\
\hline
\makecell{Monotonic\\function} & $P\subseteq \REAL^d$ & \makecell{Monotonic\\function\\$f':\REAL\to[0,\infty)$} & $\max_{p\in P}f'(p)$ & $\makecell{C\subseteq P\\ |C|=2}$ & Unweighted & $O(n)$ & $O(d)$\\
\hline
\makecell{Deviation\\mean (1)} & \makecell{$P\subseteq \REAL^d$\\$w:P\to \REAL$\\$\sum_{p\in P}w(p)=1$} & $x\in\REAL^d$ & $\sum_{p\in P}w(p)(p-x)$ & $\makecell{C\not\subseteq P\\ |C|=1}$ & Unweighted & $O(n)$ & $O(d)$\\
\hline
\makecell{Deviation\\mean (2)} & \makecell{$P\subseteq \REAL^d$\\$w:P\to \REAL$} & $x\in\REAL^d$ & $\sum_{p\in P}w(p)(p-x)$ & $\makecell{C\subseteq P\\ |C|=d+1}$ & $u:C\to \REAL$ & $O(nd^3)$ & $O(d^2)$\\
\hline
\makecell{Deviation\\mean (3)} & \makecell{$P\subseteq \REAL^d$\\$w:P\to \REAL$\\$\sum_{p\in P}w(p)=1$} & $x\in\REAL^d$ & $\sum_{p\in P}w(p)(p-x)$ & $\makecell{C\subseteq P\\ |C|=d+2}$ & \makecell{$u:C\to [0,1]$\\$\sum_{p\in C}u(p) = 1$} & $O(nd^3)$ & $O(d^2)$\\
\hline
\makecell{$1$-mean (1)} & \makecell{$P\subseteq \REAL^d$\\$w:P\to \REAL$} & $x\in\REAL^d$ & $\sum_{p\in P}w(p)\norm{p-x}^2$ & $\makecell{C \not\subseteq P\\|C|=2}$ & \makecell{Unweighted} & $O(n)$ & $O(d)$\\
\hline
\makecell{$1$-mean (2)} & \makecell{$P\subseteq \REAL^d$\\$w:P\to \REAL$} & $x\in\REAL^d$ & $\sum_{p\in P}w(p)\norm{p-x}^2$ & $\makecell{C \subseteq P\\|C|=d+2}$ & \makecell{$u:C\to\REAL$} & $O(nd^3)$ & $O(d^2)$\\
\hline
\makecell{$1$-mean (3)} & \makecell{$P\subseteq \REAL^d$\\$w:P\to \REAL$\\$\sum_{p\in P}w(p)=1$} & $x\in\REAL^d$ & $\sum_{p\in P}w(p)\norm{p-x}^2$ & $\makecell{C \subseteq P\\|C|=d+3}$ & \makecell{$u:C\to [0,1]$\\ $\sum_{p\in C}u(p) = 1$} & $O(nd^3)$ & $O(d^2)$\\
\hline
\makecell{Matrix $\ell_2$ norm} & \makecell{$P\subseteq \REAL^d$} & $x\in\REAL^d$ & $\sum_{p\in P}(p^Tx)^2$ & $\makecell{C \subseteq P\\|C|=d^2+1}$ & \makecell{Unweighted} & \makecell{$O(nd^2)$} & $O(d^3)$\\
\hline
\makecell{Linear regression} & \makecell{$P\subseteq \REAL^d\times \REAL$} & $x\in\REAL^d$ & $\sum_{(p,b)\in P}(p^Tx-b)^2$ & $\makecell{C \subseteq P\\|C|=d^2+1}$ & \makecell{Unweighted} & \makecell{$O(nd^2$)} & $O(d^3)$\\
\hline
\end{tabular}
\end{adjustbox}
\caption{\it \changed{Example of simple coresets that can be used to compute a cost function exactly for an input set $P$ of $n$ points, with no approximation error.  The columns represent, respectively, the name of the related problem, the type of input (possibly weighted via a weight function $w$), the family of possible queries that can be answered via the coreset, the cost function that the coreset should be used to evaluate, the properties of the coreset (and its weight vector $u$), property of the coreset weights, the construction time of the coreset, and the time it takes to answer a specific query using the coreset. }\label{table:acc}
}
\end{table}

\begin{table}[ht]
\centering
\begin{tabular}{|c|c|c|c|}
  \hline
  Problem & Size & Paper \\
  \hline
  Euclidean $k$-means & $O(k \epsilon^{-d} \log^{d+2} n)$ & \cite{sariela} \\
  Euclidean $k$-means & $O(k^3 \epsilon^{-(d+1)} \log^{d+2} n)$ &  \cite{sarielb}  \\
  Euclidean $k$-means & $O(d k^2 \epsilon^{-2} \log^8 n)$ & \cite{chen2009coresets} \\
  Euclidean $k$-means & $O(d k \log k \epsilon^{-4} \log^5 n)$ & \cite{FL11} \\
  Euclidean $k$-means & $\tilde{O}((d/\epsilon)^{O(d)}  k \log^{O(d)} n)$ &  \cite{Ackermann2012}\\

  Metric $k$-means & $O(\epsilon^{-2} k^2 \log^8 n)$ & \cite{Chen} \\
  Metric $k$-means & $O(\epsilon^{-4} k \log k \log^6 n)$  & \cite{FL11} \\

  Euclidean $k$-median&  $O(d k^2 \epsilon^{-2} \log^8 n)$ & \cite{chen2009coresets} \\
  Euclidean $k$-median &  $O(k \epsilon^{-d} \log^{d+2} n)$ & \cite{sariela} \\
  Euclidean $k$-median &  $O(k^2 \epsilon^{-O(d)} \log^{d+1} n)$ &  \cite{sarielb}  \\
  Euclidean $k$-median &  $O(d \epsilon^{-2} k \log k \log^3 n)$  & \cite{FL11} \\

  Metric $k$-median &  $O(k^2 \epsilon^{-2} \log^8 n)$ & \cite{chen2009coresets} \\
  Metric $k$-median  & $O(\epsilon^{-2} k \log k \log^4 n)$  & \cite{FL11} \\

  Euclidean $M$-estimators  & $O(\epsilon^{-2} k^{O(k)} d^2 \log^5 n)$& \cite{feldman2012data} \\

  Metric $M$-estimators  & $O(\epsilon^{-2} k^{O(k)} \log^7 n)$ & \cite{feldman2012data} \\
  Metric $M$-estimators   & $O(\epsilon^{-2} k \log k \log n)$ & \cite{braverman2016new} \\
  \hline
\end{tabular}
\caption{\it \changed{Coresets for the problem of $k$-clustering (points to points). The input is a set of $n$ points in $\REAL^d$ (Euclidean space) or any other metric space. Squared and non-squared distances are denoted by $k$-means and $k$-median respectively. $M$-estimators also handle outliers that ignore farthest distance by taking e.g. $\min\br{dist(p,q),1 }$ as the distance between two points.}
\label{table:k}}
\end{table}

\section{Problem types\label{sec:prob}}
\paragraph{The input set $P$ }usually consists of finite number of $n=\sum_{p\in P}w(p)$ points in some metric or Euclidean space. Some coresets have size independent of $n$, and may be applied on a set $P$ that consists of infinite number of points, e.g., an analog and continuous signal or image~\cite{FMS07,HK05,feldman2012effective,rosman2014coresets}. Every point may be assigned a positive multiplicative weight, which usually does not make the problem easier or harder. This is necessary when computing coreset for coresets as in the case of composable coresets.

\paragraph{The fitting error $f$} is usually a Log-Lipschitz function of some distance function and thus satisfies the weak triangle inequality in some sense, i.e., $\dist(p,x)-\dist(p',x)$ is \ariel{bounded} by the distance from $p$ to $p'$ for every $p,p'\in P$ and $x\in X$, up to a small multiplicative factor. This includes a distance to the power of a constant $z\geq 0$ or M-estimators that ignore large distances and thus handle outliers. When the query $X$ is a set of centers as in $k$-means, or a shape as in low-rank approximations, or $k$-line means, we usually define $\dist(p,X)=\min_{x\in X}\dist(p,x)$; see~\cite{schulman,braverman2016new}.

\paragraph{Projective clustering }is one of the fundamental type of problems that is suitable for coresets~\cite{aga1,santosh,HV02,ProcAga02,feldman2013turning}. Here, each item in the set of queries $\mathcal{X}$ is a set of $k$ affine $j$-dimensional subspaces. Special cases include $k$-means/median/center $(j = 0)$ in a metric or pseudo-metric spaces (discrete versions)~\cite{CG05,CGTS02}, low-rank approximation (PCA) where $k = 1$~\cite{CGTS02,DrineasMM06}, or well conditioned basis~\cite{DasguptaDHKM08} for different distance functions. Many other problems can be reduced to these kind of problems via linearizations in high-dimensional space.
Projective clustering is a special case of Dictionary learning as explained in~\cite{ffs11}.

\paragraph{Supervised learning }is done recently using coresets, where each input point $(p,y)$ includes a discrete or continuous label. E.g. for learning kernels, $y\in\br{-1,1}$, and the distance function is $f((p,y),x)=y\cdot \phi(p\cdot x)$ for some kernel function $\phi:\REAL\to[0,\infty)$. Unlike other optimization techniques, if the coreset is composable we may compute a coreset for each class $y$ independently and then return the union of coresets as the final coreset~\cite{tolochinsky2018coresets,joshi2011comparing,loffler2009shape}.

\paragraph{Generalization error }is less relevant for most of the coresets. Unlike in machine learning and more like in Computer Science and Computational Geometry, there is usually no assumption that the data is a set of i.i.d samples from some known or unknown distribution.
The question of what problem to solve, using which model and how to avoid overfilling is related to techniques such as maximum-likelihood or Empirical Risk-Minimization. Exceptions include Model Selection in Section~\ref{sec:why}.

\section{Generic Coreset Construction\label{sec:generic}}
Many of the existing and especially old coreset constructions can be simplified, improved and
have better guarantees using modern and retrospective analysis. In this section we try to
give a generic algorithm that can be used to generate many of these existing coresets, which continues and improve such tries from~\cite{FL11,braverman2016new}.

\paragraph{Reduction to off-line construction on a pair of coresets }is allowed, assuming composable coresets (strong or weak) construction which support the streaming model as explained in Section~\ref{sec:what}. Using the merge-and-reduce tree (see there), we can assume that we are given a set of $m \ll n$ points, which is a union of two other coresets, and the goal is to reduce it by half to a coreset of size $m/2$, for the smallest possible value of $m$. This assumption usually adds extra polynomial factors in $\log (n)$, since $\eps$ is replaced with $\eps/(\log n)$ due to the increasing error over the levels of the tree.

\paragraph{$\eps$-Sample. }Many, if not most, of the existing coresets can be defined formally as an $\eps$-sample,
which is a generalization of the definition of $\eps$-nets in computational geometry for binary functions~\cite{LLS01,ES04,HauWel86}. Given a query space $(P,w,\mathcal{X},f)$ and an error (usually constant) $\eps\in(0,1)$, an $\eps$-sample is a query space $(C,u,\mathcal{X},f)$ where $C\subseteq P$, such that $\cost((w(p)f(p,x))_{p\in p})=\sum_{p\in P}w(p)f(p,x)$ is approximated by $\cost(u(q)(f(q,x))_{q\in C})$ for every query $x\in\mathcal{X}$, up to an additive error of $\eps$, i.e.,
\[
\left| \sum_{p\in P}w(p)f(p,x)-\sum_{q\in C}u(p)f(q,x)  \right|\leq \eps.
\]
Hence, $C$ is a weighted subset strong coreset, and is also composable coreset as defined in Section~\ref{sec:what}.

For example, the common $(1\pm \eps)$-multiplicative approximation error for $\sum_{p\in P}w(p)g(p,x)$ can be obtained by defining $f(p,x)=g(p,x)/\sum_{q\in P}w(q)g(q,x)$. Smaller bound on the error as in Grid coresets may be obtained by defining $f(p,x)=g(p,x)/\sum_{q\in P}w(q)g(q,x^*)$, where $x^*$ is an optimal query. If this is hard or impossible, larger (no-longer multiplicative) error of $\Delta(P,x)\geq 0$ can be obtained by defining $f(p,x)=g(p,x)/\Delta(P,x)$.

\paragraph{Simple $\eps$-sample. }If the input weights are non-negative, i.e., $w : P\to[0,\infty)$, then PAC-learning~\cite{LLS01,BEHW89,H90}, which generalizes Hoeffding's inequality~\cite{hoeffding1963probability}, proves that a simple i.i.d uniform sampling (if $w\equiv 1/n$, or proportional to $w$ in general) yields an $\eps$-sample with probability at least $1-\delta$. The size of the sample depends polynomially on $\max_{p\in P, x\in\mathcal{X}} |f(p,x)|$, the sum of weights $\sum_{p\in P}w(p)$, $1/\eps$, $\log(1/\delta)$, and a measure of complexity for a query space that is related to generalized VC-dimension~\cite{FL11,braverman2016new}.

While uniform sample is the most common and generic way to compute an $\eps$-sample, generic deterministic constructions (usually of size exponential in $d$) was suggested in~\cite{Matousek95}, and obtain the same or better bounds of most of the Grid coresets, after proper scaling by the sup-weights below. See Section~\ref{sec:cons} and Fig.~\ref{figa}(right).

\paragraph{Sup-weights } were suggested in~\cite{FL11} to reduce the size or the error of the sample above by a factor of $n$.  The main observation is that we can replace the weight function $w$ by any function $m:P\to[0,\infty)$ and $g(p,x)=w(p)f(p,x)/m(p)$, we obtain the same desired cost $\sum_{p\in P}w(p)f(p,x)=\sum_{p\in P}m(p)g(p,x)$. The sample size above now depends on $\max_{p,x} |g(p,x)|$, and the new total weight $\sum_{p\in P}m(p)$.
To minimize them, we assign to each point $p\in P$ a weight $m(p)=w(p)\max_{x\in \mathcal{X}}|f(p,x)|$ (more generally, its \changed{supremum}). That is, we minimize the total weight under the constraint
\[
\max_{p,x}|g(p,x)|=\max_{p,x}\frac{w(p)|f(p,x)|}{m(p)}=1.
\]
Computing weaker bounds $m'(p)\geq m(p)$, say, up to a multiplicative constant factor for every point $p\in P$, would increase the total weight and size of resulting coreset by the same factor.

This new optimization problem is independent of $\eps$ or one of the many algorithms to construct $\eps$-sample (randomly or deterministically) on the new re-weighted set. Hence, it simplifies the analysis and results of many previous papers.

\paragraph{Sensitivity }is a special case of this reweighting, where we wish to obtain multiplicative $1\pm\eps$ approximation for a non-negative function $f$, i.e., $f(p,x)=g(p,x)/\sum_{q\in P}w(q)g(q,x)$ as explained in the beginning of this section. Here,
\[
m(p)=\max_{x}w(p)f(p,x)=\max_{x}\frac{w(p)g(p,x)}{\sum_{q\in P}w(q)g(q,x)}.
\]
See e.g.~\cite{C09,LS10,braverman2016new,NIPS}.

Finding small sup-weights or sensitivities for different cost functions is the main challenge of many current coreset papers. Sometimes it may be harder than the original optimization problem of finding $x^*$. In fact, the problems are usually related as explained below.

\paragraph{The chicken-and-egg }problem stems from the fact that the optimal query $x^*$ is usually required to bound sensitivity or sup-weights; see Fig.~\ref{figa}(right) and Section~\ref{sec:bound}. However, this is usually the main motivation
for constructing the coreset in the first place. There are a few lee-ways: first, it can be assumed that the size of the input is the size of a pair of coresets $m\ll n$ as explained in the beginning of this section. So inefficient algorithms
(say, polynomial in their input size $m$), would still yield linear construction time in $n$. However, for
NP-hard problems such as $k$-means clustering, the running time would still be exponential in $k$. For other problems (such as general projective clustering) even inefficient optimization algorithms might not be known.

\begin{figure}[ht]
\centering
\fbox{\includegraphics[scale=0.2,width=0.3\textwidth]{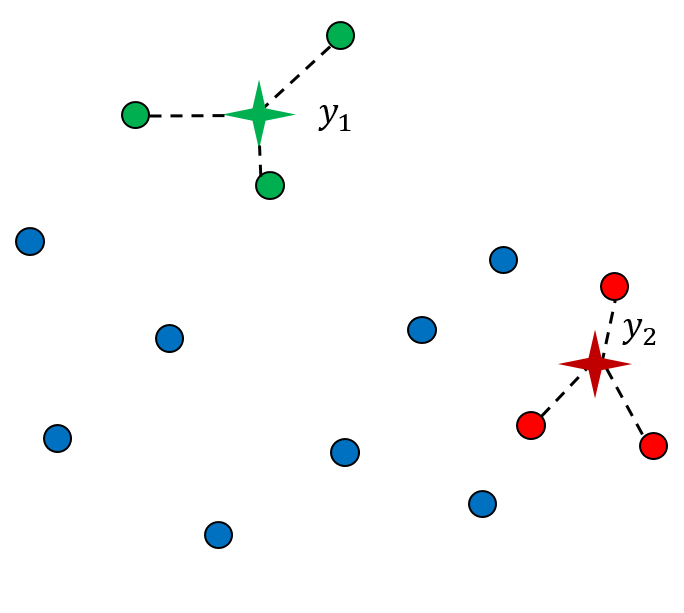}}
\vtop{\vskip-3.8cm\hbox{\fbox{\includegraphics[scale=0.2,width=0.3\textwidth]{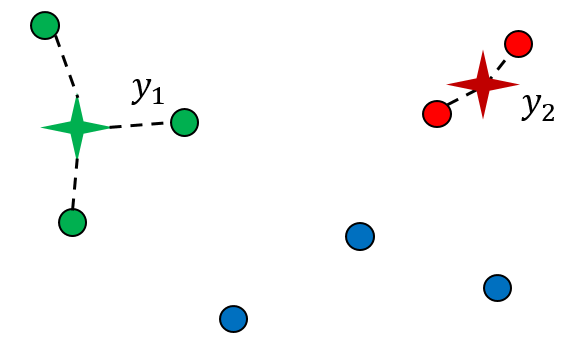}}}}
\includegraphics[height=5cm,width=0.35\textwidth]{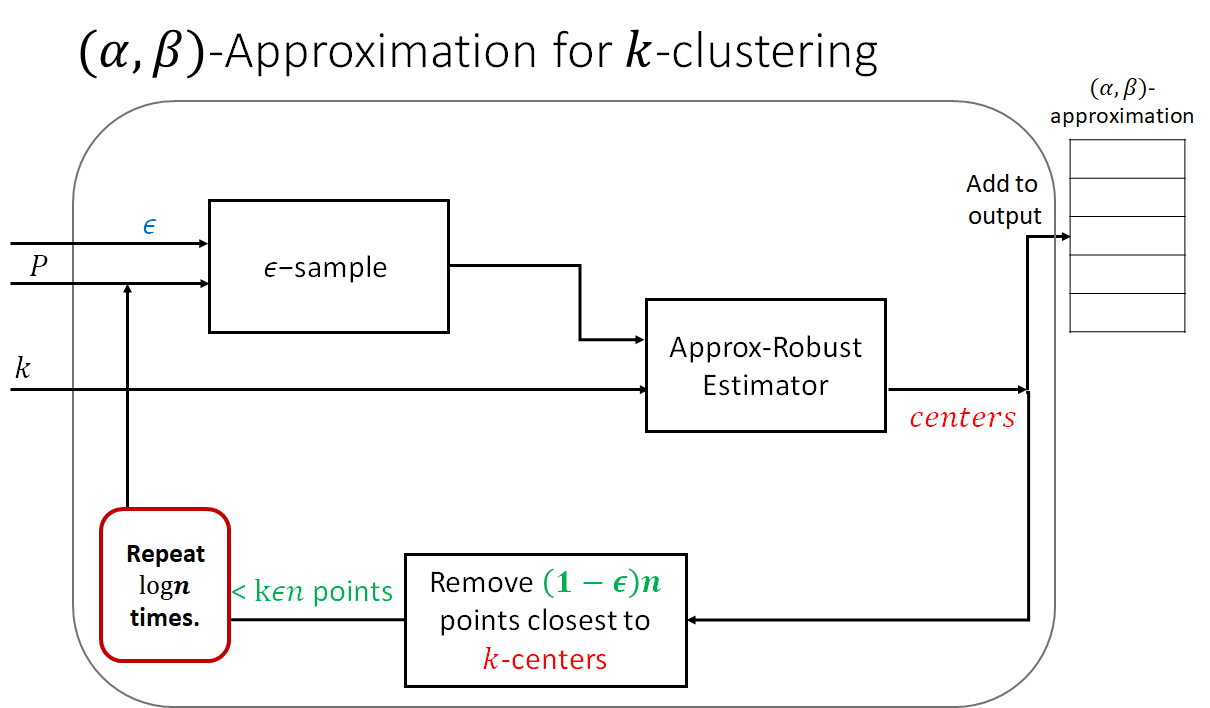}
\caption{\label{figb}\scriptsize \textbf{(left)} A robust approximation to the optimal clustering (the stars) is computed on a small $\eps n$-sample. About half of the closest points in the full set are then removed. \textbf{(middle) }The process continues recursively on the remaining half and new robust approximation (the new stars) are produced, until there are no points left. \textbf{(right)} The output $(\alpha,\beta)$-approximation is the union of robust estimators during the $O(\log n)$ iterations.}
\end{figure}

\paragraph{Bicriteria or $(\alpha,\beta)$-approximation }for the optimal solution can be used in such cases, which is a set $B \subseteq\mathcal{X}$ of $\beta \geq 1$ queries whose fitting cost $\sum_{p\in P}\min_{x\in B}f(p,x)$ is larger by a factor of at most $\alpha\geq 1$ compared to the cost of the optimal solution; see Fig.~\ref{figb}. The bound on the total sup-weights or sensitivities usually depends polynomially on $\alpha$ and $\beta$ so near-logarithmic bounds for $\alpha,\beta$ are reasonable. 

\paragraph{Generic algorithm for $(\alpha,\beta)$-approximation }is suggested in~\cite{FL11} following~\cite{FFSS07}.
First we compute an $(\alpha,\beta,1/4)$-approximation $X\subseteq\mathcal{X}$ to the \emph{robust optimal} query as explained below, then we compute for each point $p\in P$ its distance $\min_{x\in X}f(p,x)$ to its closest center in $X$, and remove half of the closest points (or weights, in general). We continue recursively on the remaining $n/2$ input points (or total weight) until there are no points left; see Fig.~\ref{figb}. The running time is only linear in the number of input points due to the geometric sequence of points in each iteration.

Unlike techniques such as RANSAC~\cite{choi1997performance}, while the first iteration uses random sampling, the
overall algorithm is adaptive: in each layer we remove the ``main stream/cluster" of the data.
``Hidden isolated" clusters may not be caught by uniform sampling in the first iteration but
would be left and be discovered during the last iterations.

\paragraph{Robust optimal} query $x\in\mathcal{X}$ minimizes the sum of distances to its closest (say, half) input points $P_{x,1/2}$, i.e., ignoring a constant fraction ($1/2$) of outliers. Computing such a query is usually much harder than computing the optimal query $x^*$. However, computing an approximation $\tilde{x}$ that serves only quarter of the points, such that $\sum_{p\in P_{\tilde{x},1/4}}f(p,\tilde{x})\leq \sum_{p\in P_{x,1/2}}f(p,x)$ is much easier, since $\eps$-sample (in particular, uniform sample) is a coreset for this problem~\cite{FL11}. An $(\alpha,\beta,1/4)$ approximation allows multiplicative factor $\alpha$ of error to the last cost, and using $|X|=\beta$ centers.

For example, for problems such as $k$-means/median/center, an $\eps$-sample $S$ for $k$ balls, i.e., uniform sample of size $|S|=O(k)$ is also the \changed{desired} $(1,|S|, 1/4)$-robust estimator. Generic algorithm for computing such robust estimator can be found in~\cite{FL11}. In practice, it may be computed via heuristics such as EM-estimators, but without provable bounds~\cite{NIPS,mclachlan2007algorithm}.

\paragraph{Bootstrapping }is used to reduce the overall size of the $\eps$-coreset after computing an $\eps$-sample that is based on sup-weights that are in turn based on an $(\alpha,\beta)$-approximation. These total sup-weights or sensitivities usually depend polynomially on $\alpha$ and $\beta$, and so does the final coreset. To remove these factors we compute the coreset (off-line, on each subset of the merge-and-reduce tree) using, say,  $\eps'=1/2$ to obtain a small coreset that can be used to compute a constant factor approximation to the optimal query ($\alpha'=\beta'=O(1)$). We then compute the coreset for the desired $\eps\in(0,1)$ using this $(\alpha',\beta')$-approximation instead of the previous $(\alpha,\beta)$-approximation to obtain coreset that depends on $\alpha'=\beta'=O(1)$. See e.g.~\cite{sariela}.

In this sense, the framework in this section can be seen as a series of improved approximations, from initial $(\eps n)$-samples that are based on uniform samples, to sup-weights sampling that are based on a rough $(\alpha,\beta)$-approximation that is based in turn based on robust estimators in each iteration, which are eventually replaced by $(\alpha',\beta')$-approximation and smaller total sup-weights/sensitivity. This process is then applied on each small subset (node) of the merge-reduce tree.

\section{Bounding Total Sup-weights\label{sec:bound} and Dimension}
In this section we suggest generic existing technique to bound sup-weights and dimension of query space.

\subsection{Bounding sup-weight}
Most of the techniques for bounding total sensitivity or sup-weights in general are based on
a given $(\alpha,\beta)$-approximation for the optimal query of the corresponding query space, as explained in the previous
section. For simplicity, we assume in this section that we are given the optimal solution $(\alpha=\beta=1)$.
Otherwise we use the bootstrapping technique above.

\paragraph{Grids. }First coresets discretized/clustered the space into a grids of cells around the optimal
solution as explained in Section~\ref{sec:query} and Fig.~\ref{figa}(right), such that every input point $p$ in a grid cell $\square$ has the same distance to every query up to an additive factor of $\eps f(p,x^*)$. Hence, the sup-weight is
\[
\frac{f(p,x)}{f(P,x^*)}
\leq \frac{f(p,x)}{f(P\cap \square,x^*)}
\leq  \frac{1}{|P\cap \square|}.
\]
The sum of the last term over every point $p\in P\cap \square$ in the square is $1$, so the total sup-weights is the number of cells in the grid, which is usually exponential in $d$. This is why deterministic $\eps$-sample constructions, whose time is exponential in $d$, are used in these coreset constructions. The approximation error for $f(P,x)$ is $\eps f(P,x^*)\leq \eps f(P,x)$ as explained in Section~\ref{sec:generic}.

\paragraph{Projection on the optimal query }allows us to reduce the problem to a problem on a
smaller dimension which in turn reduces total sup-weights. We first add the projected input $P'$ to the coreset, and then compute coreset for the difference $f(P,x)-f(P',x)\leq f(P,P')$, by bounding the sup-weight of $f(p,x)-f(p',x)\leq f(p,p')$ using the weak triangle inequality. Here we assume that $f(P,P')$ and $f(p,p')$ are well defined.
The resulting coreset is the union of $P'$ with a weighted subset of pairs that can be replaced by two points: $p$ with positive weight and $p'$ with a negative weight, as explained in~\cite{feldman2010coresets,feldman2012single}. For problems such as $k$-means we have that $P'$ consists of only $k$ points, and the negative weights can actually be removed~\cite{FL11,braverman2016new}. For other problems we need to compute a coreset for $P'$ of size $|P|=n$ but whose dimension is smaller. See following techniques.

\paragraph{Bounding sensitivity }directly using the previous technique is possible since the sensitivity of a point is bounded by $f(p,x^*)/f(P,x^*)$ plus the sensitivity of its projected point $p'$ above with respect to the set $P'$, as proven in~\cite{varadarajan}. The coreset is usually larger compared to previous technique (e.g. for $k$-means the total sensitivity is $k$ and not $1$). However, the resulting coreset is a weighted subset, with only positive weights.

\begin{figure}[ht]
\centering
\includegraphics[height=5cm, width=0.3\textwidth]{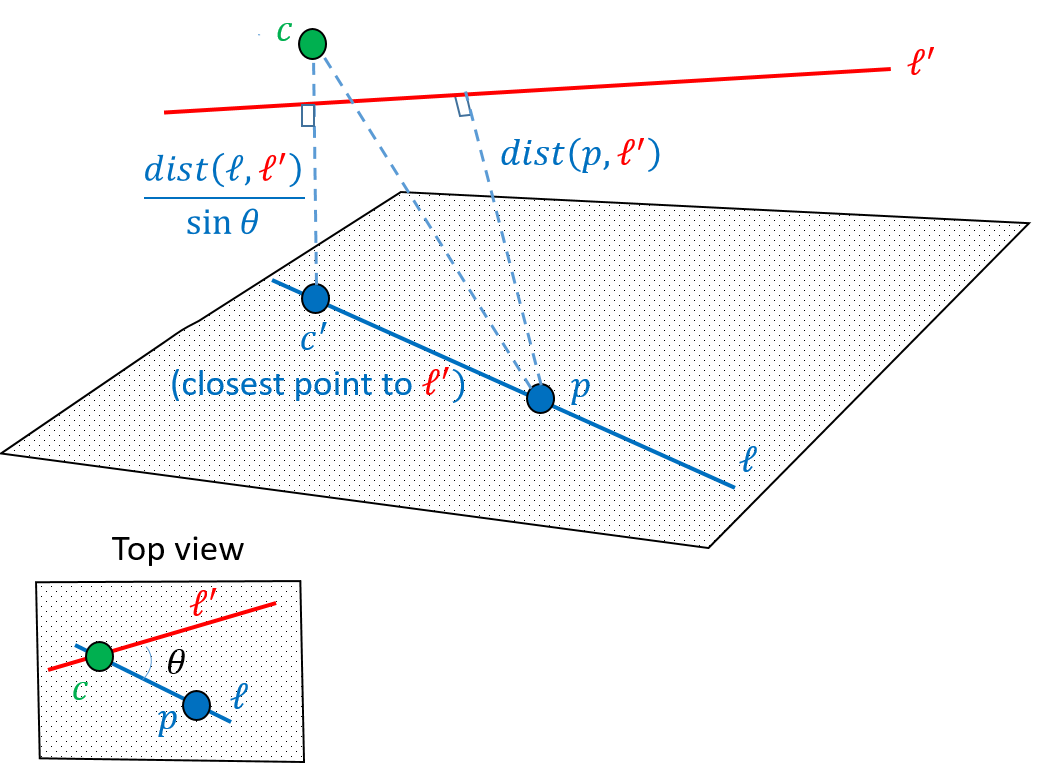}
\includegraphics[scale=0.8,width=0.3\textwidth]{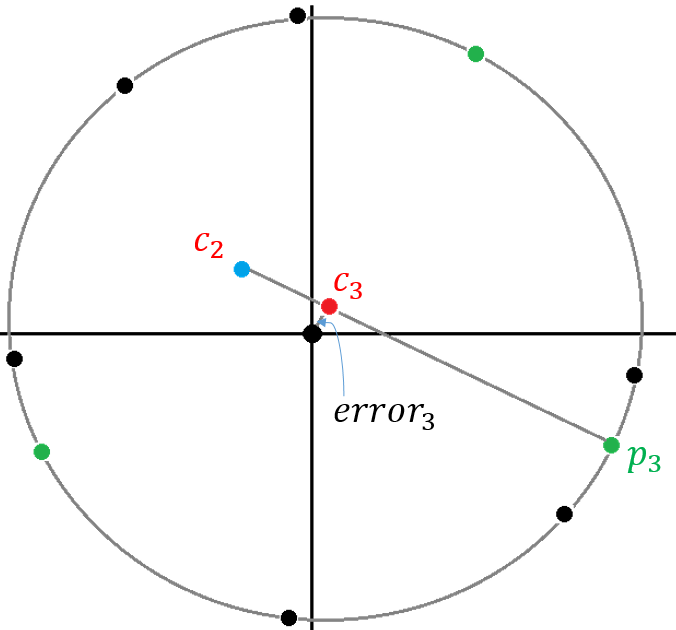}
\caption{\label{figc}\scriptsize \textbf{(left) }For every pair of lines $\ell$ (in blue) and $\ell'$ (in red) there is a center $c$ (in green) and $w\geq 0$ such that the distance from every $p\in\ell$ to $\ell'$ is the same as its distance to $c$ multiplied by $w$.  \textbf{(middle) }An arbitrary input point $c_1$ is the first center and coreset point. Its farthest input point $p_2$ is the second coreset point, and $c_2$ is the closest point between them to the origin.  \textbf{(right) } The next coreset point $p_3$ is the farthest input point from $c_2$, and $c_3$ is the new closest point to the origin. After $i=1/\eps$ iterations we have $\text{error}_i=\norm{c_i}\leq \eps$.}
\end{figure}

\paragraph{The cUTE decomposition }is used to bound sensitivity of higher-dimensional centers
than points, such as in projective clustering where $j\geq 1$. Using the above techniques we can reduce the dimension of $P$ to the dimension $j$ of the optimal query by replacing it with $P'$. It was proved in~\cite{feldman2010coresets,FFS06} that in this case every $m$-dimensional affine subspace $X$ can be replaced by an affine $(j-1)$-dimensional subspace $X'$ and a constant $c > 0$ such that the distance from every point in $P'$ to $X$ is the same as the
distance to $X'$ multiplied by a constant (weight) $c$. See Fig.~\ref{figb}(left) for the case $j=1$. This is a special case for the claim that there is a factorization $A=cU^TE$ for every $A\in\REAL^{m\times j}$ and $E\in\REAL^{d\times j}$, such that $U^TU=I$ and $c\geq 0$.

The reduction is from a set $P$ in $\REAL^d$ and $j$-dimensional queries, to a set $P'$ in $\REAL^j$ with $j$-dimensional queries. We can then apply this reduction recursively $j-1$ times till the centers (queries) are points. Unfortunately, careful analysis shows that the final total sensitivity is exponential in $j$~\cite{feldman2010coresets}.
%
%

\paragraph{Weighted centers }(multiplicative weight $w'(x)$ for every query $x$) occur: (i) when clustering to $k\geq 2$
(multiple) subspaces, which reduce to handling weighted $(j-1)$-subspaces as explained above, and (ii) for clustering weighted facilities (centers), e.g. when the travelling costs to a center
at the sea/air and a center of the same distance on the ground are not the same, and (iii)
handling $k$-clustering with $m$ outliers (centers whose weight is infinity), or $M$-estimators $f'(p,x)=\min\br{f(p,x),c)}$ for some constant threshold $c>0$.

Bounds on the sensitivities for these cases are usually exponential in the number of centers and also depends on $\log(n)$; See Section~\ref{sec:cons}.

\paragraph{Reduction of $\cost(\cdot)$ from $\norm{\cdot}_1$ to $\norm{\cdot}_{\infty}$}. Suppose that we have a subset coreset construction for our query space, $\eps=1/2$, and any input $P$ of $n$ points,  where $\cost(\cdot)=\norm{\cdot}_{\infty}$, i.e., covering queries. If the output coreset consists of at most $m$ points, then it was proven in~\cite{FSS13} based on a simple generalization of~\cite{VX12-soda} (for projective clustering), that the total sensitivity is bounded by $O(m\log n)$.

\paragraph{Beyond sup-weights. }The max-sampling framework is very generic but for specific problems we might get better coreset constructions. This includes pre-processing techniques such as Dimension Reduction before computing the coreset~\cite{FSS13}. The sparsity of the input is loss in this case, so recursive partitioning of the data can be used instead~\cite{BF16}.
Convex optimization techniques such as the Frank-Wolfe algorithm~\cite{BC08,CL10} produces a sparse solution for a
given function which are not composable, as explained in Section~\ref{sec:query}. However, a coreset construction may be formulated as such a convex optimization problem where we wish to compute a sparse distribution over the input (the set of coreset weights) that satisfies some requirements. See for example deterministic small coresets for 1-mean and low-rank
approximation~\cite{FeldmanOR17,feldmanmik}.

\subsection{Bounding Queries Dimension}
The dimension of a query space is the VC-dimension of the corresponding range space of
the $\eps$-sample as explained in Section~\ref{sec:generic}. There are many techniques for bounding the VC-dimension of such a range space; see e.g.~\cite{anthony-bartlett:1999a}. The range space is the same for the same pseudo-distance function to any power of $z\geq1$. For example, $k$-median/median/center queries has the same corresponding range space: the sets of $k$ balls in $\REAL^d$. To bound the VC-dimension it is usually useful to consider the squared distances, which are polynomial functions in case of points in $\REAL^d$, as in the case of projective clustering. The VC-dimension for such $n$ polynomial functions is their degree $d$~\cite{FSS13}. More generally, if we can answer a query in $O(d)$ arithmetic operations and exponential functions then the corresponding dimension of the query space is also polynomial in $d$.

\paragraph{Weak composable corests }maintains the approximated optimal solution using smaller sample, by replacing the VC-dimension by Generalized VC-dimension that may be smaller by order of magnitudes, or even independent, on parameters such as dimension $d$; see Section~\ref{sec:query}.

\paragraph{Homomorphic query space }is a query space whose dimension is smaller, but sufficient to
answer every query of a query space of a larger dimension. For example, squared distances
to any shape that is contained in a $k$-subspace can be approximated up to factor of $1\pm\eps$ by another rotated shape that is contained in a fixed $(k/\eps)$-dimensional subspace~\cite{FSS13}.

\paragraph{Deterministic Constructions }for $\eps$-sample might be smaller than the non-tight bound on the size that is obtained from uniform random sampling or other techniques. See Fig.~\ref{figc}(right) and Section~\ref{sec:cons}.

\subsection{Lower bounds\label{sec:lb}}
\changed{Since there is no exact definition for a coreset, the most general definition is ``any data structure
that can answers every query in the query set". Here, the computation issues from Section~\ref{five} are ignored. The size of the coreset can be measured by the total number of the bits it takes to store it in memory. A lower bound under this assumption is usually proved via communication protocols where Alice tries to answer a query based on an input data that Bob has, using minimum communication between
them~\cite{muthukrishnan2005data,munteanu2018coresets}.

Other lower bounds are known for specific problems and coreset types, such as for the size of weighted subset coresets~\cite{H04,HP06,edwards2005no}, dependency on the stretch ratio of the input~\cite{VX12-soda}, or order of input points~\cite{AssadiK17}. For random constructions, lower bounds may be computed using e.g. the coupon collector~\cite{braverman2016new}, and smaller deterministic versions are possible via~\cite{batson2014twice}.
}

\section{Future Research\label{future}}

Another direction is to apply coresets for problems beyond data reductions, as the following two examples.

\paragraph{Private coresets.   }
In the recent years, we started to solve theoretical and practical open problems using coresets
in fields that seemingly have nothing to do with data reduction. For example, in~\cite{feldman2009private,feldman2017coresets} \emph{private coresets} that preserve differential privacy were suggested to enable answer unbounded
number of queries with no leakage of privacy for a specific user. It was also proved there
that a small coreset for \emph{any} problem implies a private coreset that introduces a small additive noise, for the corresponding query space. However, the proof is not constructive. Having a constructive proof will turn the existing numerous coreset constructions into private coresets for many open problems in machine learning. Such a contribution is especially important since lack of practical results for machine learning is one of the critique on differential privacy~\cite{bambauer2013fool}. A more humble but important result is to suggest private coresets for specific problems, based on their corresponding non-private versions as in~\cite{feldman2009private,feldman2017coresets}.

\paragraph{Fully Homomorphic Encryption (FHE).} These algorithms get encrypted data $[[P]]$ of $P$ on the server (e.g. cloud) and output the encrypted result $[[X]]$ of $X$ without revealing the secret encryption/decryption key. However, FHE algorithms are usually impractically slow. Computing coresets using FHE might be easier and faster as was recently shown in~\cite{AkaviaFS18,ASF19}. The main idea is that instead of solving the complete problem on the server's side, only the coreset will be computed on the server and transmitted (encrypted) to the client. The client will then extract locally the desired result from the coreset in short time due to the small size of the coreset. It turns out that for many problems the task of computing a coreset in the FHE computation model is significantly easier than solving the original problem. This is similar to the fact that we can compute coresets in linear or near-linear time for NP-hard problems such as approximating $k$-means up to a small factor of $1\pm\eps$. Sketch matrices are particular useful for FHE since they involve only multiplication by a matrix, which is an operation that can be implemented very efficiently via FHE; see~\cite{AkaviaFS18,ASF19} for details.

\paragraph{Deterministic coresets. }Using the sup-sampling approach in this survey, the problem of computing coresets can be reduce to the problem of computing $\eps$-samples, after proper weighting by the sup or sensitivity of the function at hand. Computing such $\eps$-samples deterministically of size polynomial in the VC-dimension is an open problem even for the case where the query is the set of half-spaces in $\REAL^d$.

\paragraph{Deep Learning. }Coresets for Deep Learning are natural since Deep Learning usually applied on very big data-sets, in parallel, and the training time is long. One goal may be to reduce the training data for shorter training time, and another goal may be to sparsity (compress) the network itself for faster classification. Since the functions in deep learning are much more complicated than the functions in this survey, a natural approach is to compute coreset for each neuron and then train the network neuron-by-neuron as in~\cite{tolochinsky2018coresets}. Similarly, we can compress the network neuron by neuron via sensitivities of edges as was suggested in~\cite{sener2017active,cenk}. Since no coresets are known even for activation functions of a single neuron, there are many open problems also in this field.

\paragraph{Coresets for other machine learning problems. }There are many problems in traditional machine learning with no coresets, including e.g. decision trees/forest with all their variants, as well as supported vector machines and other optimization functions such as RELU.

\paragraph{Signal processing. }Coresets for fast computations of FFT or wavelets are natural, and it seems that even provable linear-time approximation algorithm for these problems are not known.

\changed{\paragraph{Implementation. }Coresets were born in theoretical computer science, and computational geometry in particular. As such, most of the results have no related software or experimental results. In fact, even pseudo code is usually missing and the algorithm is hidden in the proof. In the recent years more researchers implement coreset and open code can be found e.g. in~\cite{code1,code2,code3,code4,code5}.

\paragraph{Unified framework. }The tedious approach of computing a coreset for each problem, paper after paper, suggests to focus on holistic approaches for constructing coresets following Section~\ref{sec:generic}. This includes connection to other data summarization techniques such as sketches, dictionary learning, kernalization, compressed sensing, sparse optimization and graph sparsification.}

\newcommand{\Proc}{Proceedings of the\ } 
\newcommand{\proc}{Proceedings of the\ } 
\newcommand{\STOC}{ACM Symposium on the Theory of Computing (STOC)}
\newcommand{\stoc}{ACM Symposium on the Theory of Computing (STOC)}
\newcommand{\FOCS}{Annual IEEE Symposium on Foundations of Computer Science (FOCS)}
\newcommand{\SODA}{ACM-SIAM Symposium on Discrete Algorithms (SODA)}
\newcommand{\soda}{ACM-SIAM Symposium on Discrete Algorithms (SODA)}
\newcommand{\SoCG}{ACM Symposium on Computational Geometry (SoCG)}
\newcommand{\socg}{ACM Symposium on Computational Geometry (SoCG)}
\newcommand{\KDD}{ACM SIGKDD Conference on Knowledge Discovery and Data Mining (KDD)}
\newcommand{\CIKM}{ACM CIKM International Conference on Information and Knowledge Management (CIKM)}
\newcommand{\COLT}{Conference on Learning Theory (COLT)}
\newcommand{\PAMI}{IEEE Transactions on Pattern Analysis and Machine Intelligence (PAMI)}
\newcommand{\APPROX}{International Workshop on Approximation Algorithms for Combinatorial Optimization Problems (APPROX)}
\newcommand{\RANDOM}{International Workshop on Randomization and Computation (RANDOM)}
\newcommand{\PODS}{ACM SIGMOD-SIGACT-SIGART Symposium on Principles of Database Systems (PODS)}
\newcommand{\WAOA}{Workshop on Approximation and Online Algorithms (WAOA)}
\newcommand{\ALENEX}{Workshop on Algorithm Engineering and Experiments (ALENEX)}
\newcommand{\ALT}{International Workshop on Algorithmic Learning Theory (ALT)}
\newcommand{\alt}{International Workshop on Algorithmic Learning Theory (ALT)}
\newcommand{\VLDB}{International Conference on Very Large Data Bases (VLDB)}
\newcommand{\JComSystSci}{Journal of Computer and System Sciences}
\newcommand{\NIPS}{Annual Conference on Neural Information Processing Systems (NIPS)}
\newcommand{\ESA}{Annual European Symposium on Algorithms (ESA)}
\newcommand{\WALCOM}{Workshop on Algorithms and Computation (WALCOM)}
\newcommand{\PAKDD}{Pacific-Asia Conference on Knowledge Discovery and Data Mining (PAKDD)}
\newcommand{\ICDM}{IEEE International Conference on Data Mining (ICDM)}
\newcommand{\CSSE}{International Conference on Computer Science and Software Engineering (CSSE)}
\newcommand{\ICCV}{ International Conference on Computer Vision (ICCV)}
\newcommand{\FSTTCS}{IARCS Annual Conference on Foundations of Software Technology and Theoretical Computer Science
(FSTTCS)}
\newcommand{\fsttcs}{IARCS Annual Conference on Foundations of Software Technology and Theoretical Computer Science
(FSTTCS)}

\bibliography{newbibfile}

\begin{thebibliography}{}

\bibitem [\protect \citeauthoryear {%
Ackermann%
\ \protect \BOthers {.}}{%
Ackermann%
\ \protect \BOthers {.}}{%
{\protect \APACyear {2012}}%
}]{%
Ackermann2012}
\APACinsertmetastar {%
Ackermann2012}%
\begin{APACrefauthors}%
Ackermann, M\BPBI R.%
, M\"{a}rtens, M.%
, Raupach, C.%
, Swierkot, K.%
, Lammersen, C.%
\BCBL {}\ \BBA {} Sohler, C.%
\end{APACrefauthors}%
\unskip\
\newblock
\APACrefYearMonthDay{2012}{{\APACmonth{05}}}{}.
\newblock
{\BBOQ}\APACrefatitle {StreamKM++: A Clustering Algorithm for Data Streams}
  {Streamkm++: A clustering algorithm for data streams}.{\BBCQ}
\newblock
\APACjournalVolNumPages{J. Exp. Algorithmics}{17}{}{2.4:2.1--2.4:2.30}.
\newblock
\begin{APACrefURL} \url{http://doi.acm.org/10.1145/2133803.2184450}
  \end{APACrefURL}
\newblock
\begin{APACrefDOI} \doi{10.1145/2133803.2184450} \end{APACrefDOI}
\PrintBackRefs{\CurrentBib}

\bibitem [\protect \citeauthoryear {%
P.~Agarwal%
, Har-Peled%
\BCBL {}\ \BBA {} Varadarajan%
}{%
P.~Agarwal%
\ \protect \BOthers {.}}{%
{\protect \APACyear {2004}}%
}]{%
AHV04}
\APACinsertmetastar {%
AHV04}%
\begin{APACrefauthors}%
Agarwal, P.%
, Har-Peled, S.%
\BCBL {}\ \BBA {} Varadarajan, K.%
\end{APACrefauthors}%
\unskip\
\newblock
\APACrefYearMonthDay{2004}{}{}.
\newblock
{\BBOQ}\APACrefatitle {Approximating extent measures of points} {Approximating
  extent measures of points}.{\BBCQ}
\newblock
\APACjournalVolNumPages{Journal of the ACM}{51}{4}{606--635}.
\PrintBackRefs{\CurrentBib}

\bibitem [\protect \citeauthoryear {%
P.~Agarwal%
, Har-Peled%
\BCBL {}\ \BBA {} Varadarajan%
}{%
P.~Agarwal%
\ \protect \BOthers {.}}{%
{\protect \APACyear {2005}}%
}]{%
agarwal2005geometric}
\APACinsertmetastar {%
agarwal2005geometric}%
\begin{APACrefauthors}%
Agarwal, P.%
, Har-Peled, S.%
\BCBL {}\ \BBA {} Varadarajan, K.%
\end{APACrefauthors}%
\unskip\
\newblock
\APACrefYearMonthDay{2005}{}{}.
\newblock
{\BBOQ}\APACrefatitle {Geometric approximation via coresets} {Geometric
  approximation via coresets}.{\BBCQ}
\newblock
\APACjournalVolNumPages{Combinatorial and computational geometry}{52}{}{1--30}.
\PrintBackRefs{\CurrentBib}

\bibitem [\protect \citeauthoryear {%
P\BPBI K.~Agarwal%
\ \BBA {} Har-Peled%
}{%
P\BPBI K.~Agarwal%
\ \BBA {} Har-Peled%
}{%
{\protect \APACyear {2001}}%
}]{%
AHP01}
\APACinsertmetastar {%
AHP01}%
\begin{APACrefauthors}%
Agarwal, P\BPBI K.%
\BCBT {}\ \BBA {} Har-Peled, S.%
\end{APACrefauthors}%
\unskip\
\newblock
\APACrefYearMonthDay{2001}{}{}.
\newblock
{\BBOQ}\APACrefatitle {Maintaining the approximate extent measures of moving
  points} {Maintaining the approximate extent measures of moving
  points}.{\BBCQ}
\newblock
\BIn{} \APACrefbtitle {Proceedings of the 12th SODA} {Proceedings of the 12th
  soda}\ (\BPGS\ 148 -- 157).
\PrintBackRefs{\CurrentBib}

\bibitem [\protect \citeauthoryear {%
P\BPBI K.~Agarwal%
, Jones%
, Murali%
\BCBL {}\ \BBA {} Procopiuc%
}{%
P\BPBI K.~Agarwal%
, Jones%
\BCBL {}\ \protect \BOthers {.}}{%
{\protect \APACyear {2002}}%
}]{%
ProcAga02}
\APACinsertmetastar {%
ProcAga02}%
\begin{APACrefauthors}%
Agarwal, P\BPBI K.%
, Jones, M.%
, Murali, T\BPBI M.%
\BCBL {}\ \BBA {} Procopiuc, C\BPBI M.%
\end{APACrefauthors}%
\unskip\
\newblock
\APACrefYearMonthDay{2002}{}{}.
\newblock
{\BBOQ}\APACrefatitle {A {M}onte {C}arlo algorithm for fast projective
  clustering} {A {M}onte {C}arlo algorithm for fast projective
  clustering}.{\BBCQ}
\newblock
\BIn{} \APACrefbtitle {Proc. {ACM}-{SIGMOD} Int. Conf. on Management of Data}
  {Proc. {ACM}-{SIGMOD} int. conf. on management of data}\ (\BPGS\ 418--427).
\PrintBackRefs{\CurrentBib}

\bibitem [\protect \citeauthoryear {%
P\BPBI K.~Agarwal%
\ \BBA {} Mustafa%
}{%
P\BPBI K.~Agarwal%
\ \BBA {} Mustafa%
}{%
{\protect \APACyear {2004}}%
}]{%
agakmean}
\APACinsertmetastar {%
agakmean}%
\begin{APACrefauthors}%
Agarwal, P\BPBI K.%
\BCBT {}\ \BBA {} Mustafa, N\BPBI H.%
\end{APACrefauthors}%
\unskip\
\newblock
\APACrefYearMonthDay{2004}{}{}.
\newblock
{\BBOQ}\APACrefatitle {$k$-Means Projective Clustering} {$k$-means projective
  clustering}.{\BBCQ}
\newblock
\BIn{} \APACrefbtitle {Proc. 23rd {ACM} {SIGACT}-{SIGMOD}-{SIGART} Symp. on
  Principles of Database Systems ({PODS})} {Proc. 23rd {ACM}
  {SIGACT}-{SIGMOD}-{SIGART} symp. on principles of database systems ({PODS})}\
  (\BPGS\ 155--165).
\PrintBackRefs{\CurrentBib}

\bibitem [\protect \citeauthoryear {%
P\BPBI K.~Agarwal%
\ \BBA {} Procopiuc%
}{%
P\BPBI K.~Agarwal%
\ \BBA {} Procopiuc%
}{%
{\protect \APACyear {2000}}%
}]{%
aga1}
\APACinsertmetastar {%
aga1}%
\begin{APACrefauthors}%
Agarwal, P\BPBI K.%
\BCBT {}\ \BBA {} Procopiuc, C\BPBI M.%
\end{APACrefauthors}%
\unskip\
\newblock
\APACrefYearMonthDay{2000}{}{}.
\newblock
{\BBOQ}\APACrefatitle {Approximation algorithms for projective clustering}
  {Approximation algorithms for projective clustering}.{\BBCQ}
\newblock
\BIn{} \APACrefbtitle {Proc. 11th Annu. {ACM}-SIAM Symp. on Discrete Algorithms
  ({SODA})} {Proc. 11th annu. {ACM}-siam symp. on discrete algorithms
  ({SODA})}\ (\BPGS\ 538--547).
\PrintBackRefs{\CurrentBib}

\bibitem [\protect \citeauthoryear {%
P\BPBI K.~Agarwal%
\ \BBA {} Procopiuc%
}{%
P\BPBI K.~Agarwal%
\ \BBA {} Procopiuc%
}{%
{\protect \APACyear {2003}}%
}]{%
AP03}
\APACinsertmetastar {%
AP03}%
\begin{APACrefauthors}%
Agarwal, P\BPBI K.%
\BCBT {}\ \BBA {} Procopiuc, C\BPBI M.%
\end{APACrefauthors}%
\unskip\
\newblock
\APACrefYearMonthDay{2003}{}{}.
\newblock
{\BBOQ}\APACrefatitle {Approximation algorithms for projective clustering}
  {Approximation algorithms for projective clustering}.{\BBCQ}
\newblock
\APACjournalVolNumPages{Journal of Algorithms}{46}{2}{115--139}.
\PrintBackRefs{\CurrentBib}

\bibitem [\protect \citeauthoryear {%
P\BPBI K.~Agarwal%
, Procopiuc%
\BCBL {}\ \BBA {} Varadarajan%
}{%
P\BPBI K.~Agarwal%
, Procopiuc%
\BCBL {}\ \BBA {} Varadarajan%
}{%
{\protect \APACyear {2002}}%
}]{%
agarwal2002approximation}
\APACinsertmetastar {%
agarwal2002approximation}%
\begin{APACrefauthors}%
Agarwal, P\BPBI K.%
, Procopiuc, C\BPBI M.%
\BCBL {}\ \BBA {} Varadarajan, K\BPBI R.%
\end{APACrefauthors}%
\unskip\
\newblock
\APACrefYearMonthDay{2002}{}{}.
\newblock
{\BBOQ}\APACrefatitle {Approximation algorithms for k-line center}
  {Approximation algorithms for k-line center}.{\BBCQ}
\newblock
\BIn{} \APACrefbtitle {European Symposium on Algorithms} {European symposium on
  algorithms}\ (\BPGS\ 54--63).
\PrintBackRefs{\CurrentBib}

\bibitem [\protect \citeauthoryear {%
Aggarwal%
, Deshpande%
\BCBL {}\ \BBA {} Kannan%
}{%
Aggarwal%
\ \protect \BOthers {.}}{%
{\protect \APACyear {2009}}%
}]{%
ADK09}
\APACinsertmetastar {%
ADK09}%
\begin{APACrefauthors}%
Aggarwal, A.%
, Deshpande, A.%
\BCBL {}\ \BBA {} Kannan, R.%
\end{APACrefauthors}%
\unskip\
\newblock
\APACrefYearMonthDay{2009}{}{}.
\newblock
{\BBOQ}\APACrefatitle {Adaptive Sampling for $k$-Means Clustering} {Adaptive
  sampling for $k$-means clustering}.{\BBCQ}
\newblock
\BIn{} \APACrefbtitle {Proceedings of the 25th APPROX} {Proceedings of the 25th
  approx}\ (\BPG~15-28).
\PrintBackRefs{\CurrentBib}

\bibitem [\protect \citeauthoryear {%
Akavia%
, Feldman%
\BCBL {}\ \BBA {} Shaul%
}{%
Akavia%
\ \protect \BOthers {.}}{%
{\protect \APACyear {2018}}%
}]{%
AkaviaFS18}
\APACinsertmetastar {%
AkaviaFS18}%
\begin{APACrefauthors}%
Akavia, A.%
, Feldman, D.%
\BCBL {}\ \BBA {} Shaul, H.%
\end{APACrefauthors}%
\unskip\
\newblock
\APACrefYearMonthDay{2018}{}{}.
\newblock
{\BBOQ}\APACrefatitle {Secure Search on Encrypted Data via Multi-Ring Sketch}
  {Secure search on encrypted data via multi-ring sketch}.{\BBCQ}
\newblock
\BIn{} \APACrefbtitle {Proceedings of the {ACM} {SIGSAC} Conference on Computer
  and Communications Security, {CCS} 2018. See also:
  {https://arxiv.org/abs/1708.05811}} {Proceedings of the {ACM} {SIGSAC}
  conference on computer and communications security, {CCS} 2018. see also:
  {https://arxiv.org/abs/1708.05811}}\ (\BPGS\ 985--1001).
\PrintBackRefs{\CurrentBib}

\bibitem [\protect \citeauthoryear {%
Akavia%
, Feldman%
\BCBL {}\ \BBA {} Shaul%
}{%
Akavia%
\ \protect \BOthers {.}}{%
{\protect \APACyear {2019}}%
}]{%
ASF19}
\APACinsertmetastar {%
ASF19}%
\begin{APACrefauthors}%
Akavia, A.%
, Feldman, D.%
\BCBL {}\ \BBA {} Shaul, H.%
\end{APACrefauthors}%
\unskip\
\newblock
\APACrefYearMonthDay{2019}{}{}.
\newblock
{\BBOQ}\APACrefatitle {Secure Data Retrieval on the Cloud: Homomorphic
  Encryption meets Coresets} {Secure data retrieval on the cloud: Homomorphic
  encryption meets coresets}.{\BBCQ}
\newblock
\APACjournalVolNumPages{}{}{}{To appear}.
\PrintBackRefs{\CurrentBib}

\bibitem [\protect \citeauthoryear {%
Anthony%
\ \BBA {} Bartlett%
}{%
Anthony%
\ \BBA {} Bartlett%
}{%
{\protect \APACyear {1999}}%
}]{%
anthony-bartlett:1999a}
\APACinsertmetastar {%
anthony-bartlett:1999a}%
\begin{APACrefauthors}%
Anthony, M.%
\BCBT {}\ \BBA {} Bartlett, P\BPBI L.%
\end{APACrefauthors}%
\unskip\
\newblock
\APACrefYear{1999}.
\newblock
\APACrefbtitle {Neural Network Learning: Theoretical Foundations} {Neural
  network learning: Theoretical foundations}.
\newblock
\APACaddressPublisher{Cambridge, England}{Cambridge University Press}.
\PrintBackRefs{\CurrentBib}

\bibitem [\protect \citeauthoryear {%
Assadi%
\ \BBA {} Khanna%
}{%
Assadi%
\ \BBA {} Khanna%
}{%
{\protect \APACyear {2017}}%
}]{%
AssadiK17}
\APACinsertmetastar {%
AssadiK17}%
\begin{APACrefauthors}%
Assadi, S.%
\BCBT {}\ \BBA {} Khanna, S.%
\end{APACrefauthors}%
\unskip\
\newblock
\APACrefYearMonthDay{2017}{}{}.
\newblock
{\BBOQ}\APACrefatitle {Randomized Composable Coresets for Matching and Vertex
  Cover} {Randomized composable coresets for matching and vertex cover}.{\BBCQ}
\newblock
\APACjournalVolNumPages{CoRR}{abs/1705.08242}{}{}.
\newblock
\begin{APACrefURL} \url{http://arxiv.org/abs/1705.08242} \end{APACrefURL}
\PrintBackRefs{\CurrentBib}

\bibitem [\protect \citeauthoryear {%
Bachem%
, Lucic%
\BCBL {}\ \BBA {} Krause%
}{%
Bachem%
\ \protect \BOthers {.}}{%
{\protect \APACyear {2015}}%
}]{%
bachem2015coresets}
\APACinsertmetastar {%
bachem2015coresets}%
\begin{APACrefauthors}%
Bachem, O.%
, Lucic, M.%
\BCBL {}\ \BBA {} Krause, A.%
\end{APACrefauthors}%
\unskip\
\newblock
\APACrefYearMonthDay{2015}{}{}.
\newblock
{\BBOQ}\APACrefatitle {Coresets for Nonparametric Estimation—the Case of
  DP-Means} {Coresets for nonparametric estimation—the case of
  dp-means}.{\BBCQ}
\newblock
\BIn{} \APACrefbtitle {International Conference on Machine Learning (ICML).}
  {International conference on machine learning (icml).}
\PrintBackRefs{\CurrentBib}

\bibitem [\protect \citeauthoryear {%
Bambauer%
, Muralidhar%
\BCBL {}\ \BBA {} Sarathy%
}{%
Bambauer%
\ \protect \BOthers {.}}{%
{\protect \APACyear {2013}}%
}]{%
bambauer2013fool}
\APACinsertmetastar {%
bambauer2013fool}%
\begin{APACrefauthors}%
Bambauer, J.%
, Muralidhar, K.%
\BCBL {}\ \BBA {} Sarathy, R.%
\end{APACrefauthors}%
\unskip\
\newblock
\APACrefYearMonthDay{2013}{}{}.
\newblock
{\BBOQ}\APACrefatitle {Fool's gold: an illustrated critique of differential
  privacy} {Fool's gold: an illustrated critique of differential
  privacy}.{\BBCQ}
\newblock
\APACjournalVolNumPages{Vand. J. Ent. \& Tech. L.}{16}{}{701}.
\PrintBackRefs{\CurrentBib}

\bibitem [\protect \citeauthoryear {%
Barger%
\ \BBA {} Feldman%
}{%
Barger%
\ \BBA {} Feldman%
}{%
{\protect \APACyear {2016}}%
}]{%
BF16}
\APACinsertmetastar {%
BF16}%
\begin{APACrefauthors}%
Barger, A.%
\BCBT {}\ \BBA {} Feldman, D.%
\end{APACrefauthors}%
\unskip\
\newblock
\APACrefYearMonthDay{2016}{}{}.
\newblock
{\BBOQ}\APACrefatitle {{k}-Means for Streaming and Distributed Big Sparse Data}
  {{k}-means for streaming and distributed big sparse data}.{\BBCQ}
\newblock
\BIn{} \APACrefbtitle {Proc. of the 2016 SIAM International Conference on Data
  Mining (SDM'16).} {Proc. of the 2016 siam international conference on data
  mining (sdm'16).}
\PrintBackRefs{\CurrentBib}

\bibitem [\protect \citeauthoryear {%
Batson%
, Spielman%
\BCBL {}\ \BBA {} Srivastava%
}{%
Batson%
\ \protect \BOthers {.}}{%
{\protect \APACyear {2014}}%
}]{%
batson2014twice}
\APACinsertmetastar {%
batson2014twice}%
\begin{APACrefauthors}%
Batson, J.%
, Spielman, D\BPBI A.%
\BCBL {}\ \BBA {} Srivastava, N.%
\end{APACrefauthors}%
\unskip\
\newblock
\APACrefYearMonthDay{2014}{}{}.
\newblock
{\BBOQ}\APACrefatitle {Twice-ramanujan sparsifiers} {Twice-ramanujan
  sparsifiers}.{\BBCQ}
\newblock
\APACjournalVolNumPages{SIAM Review}{56}{2}{315--334}.
\PrintBackRefs{\CurrentBib}

\bibitem [\protect \citeauthoryear {%
Baykal%
, Liebenwein%
, Gilitschenski%
, Feldman%
\BCBL {}\ \BBA {} Rus%
}{%
Baykal%
\ \protect \BOthers {.}}{%
{\protect \APACyear {2019, to appear.}}%
}]{%
cenk}
\APACinsertmetastar {%
cenk}%
\begin{APACrefauthors}%
Baykal, C.%
, Liebenwein, L.%
, Gilitschenski, I.%
, Feldman, D.%
\BCBL {}\ \BBA {} Rus, D.%
\end{APACrefauthors}%
\unskip\
\newblock
\APACrefYearMonthDay{2019, to appear.}{}{}.
\newblock
{\BBOQ}\APACrefatitle {Data-Dependent Coresets for Compressing Neural Networks
  with Applications to Generalization Bounds} {Data-dependent coresets for
  compressing neural networks with applications to generalization
  bounds}.{\BBCQ}
\newblock
\BIn{} \APACrefbtitle {International Conference on Learning Representations
  (ICLR).} {International conference on learning representations (iclr).}
\PrintBackRefs{\CurrentBib}

\bibitem [\protect \citeauthoryear {%
Bentley%
\ \BBA {} Saxe%
}{%
Bentley%
\ \BBA {} Saxe%
}{%
{\protect \APACyear {1980}}%
}]{%
bent}
\APACinsertmetastar {%
bent}%
\begin{APACrefauthors}%
Bentley, J\BPBI L.%
\BCBT {}\ \BBA {} Saxe, J\BPBI B.%
\end{APACrefauthors}%
\unskip\
\newblock
\APACrefYearMonthDay{1980}{}{}.
\newblock
{\BBOQ}\APACrefatitle {Decomposable searching problems I. Static-to-dynamic
  transformation} {Decomposable searching problems i. static-to-dynamic
  transformation}.{\BBCQ}
\newblock
\APACjournalVolNumPages{Journal of Algorithms}{1}{4}{301--358}.
\PrintBackRefs{\CurrentBib}

\bibitem [\protect \citeauthoryear {%
Blumer%
, Ehrenfeucht%
, Haussler%
\BCBL {}\ \BBA {} Warmuth%
}{%
Blumer%
\ \protect \BOthers {.}}{%
{\protect \APACyear {1989}}%
}]{%
BEHW89}
\APACinsertmetastar {%
BEHW89}%
\begin{APACrefauthors}%
Blumer, A.%
, Ehrenfeucht, A.%
, Haussler, D.%
\BCBL {}\ \BBA {} Warmuth, M\BPBI K.%
\end{APACrefauthors}%
\unskip\
\newblock
\APACrefYearMonthDay{1989}{}{}.
\newblock
{\BBOQ}\APACrefatitle {Learnability and the Vapnik-Chervonenkis dimension}
  {Learnability and the vapnik-chervonenkis dimension}.{\BBCQ}
\newblock
\APACjournalVolNumPages{Journal of the {ACM}}{36}{4}{929--965}.
\PrintBackRefs{\CurrentBib}

\bibitem [\protect \citeauthoryear {%
Boutsidis%
, Zouzias%
, Mahoney%
\BCBL {}\ \BBA {} Drineas%
}{%
Boutsidis%
\ \protect \BOthers {.}}{%
{\protect \APACyear {2015}}%
}]{%
boutsidis2015randomized}
\APACinsertmetastar {%
boutsidis2015randomized}%
\begin{APACrefauthors}%
Boutsidis, C.%
, Zouzias, A.%
, Mahoney, M\BPBI W.%
\BCBL {}\ \BBA {} Drineas, P.%
\end{APACrefauthors}%
\unskip\
\newblock
\APACrefYearMonthDay{2015}{}{}.
\newblock
{\BBOQ}\APACrefatitle {Randomized dimensionality reduction for $ k $-means
  clustering} {Randomized dimensionality reduction for $ k $-means
  clustering}.{\BBCQ}
\newblock
\APACjournalVolNumPages{IEEE Transactions on Information
  Theory}{61}{2}{1045--1062}.
\PrintBackRefs{\CurrentBib}

\bibitem [\protect \citeauthoryear {%
Braverman%
, Feldman%
\BCBL {}\ \BBA {} Lang%
}{%
Braverman%
\ \protect \BOthers {.}}{%
{\protect \APACyear {2016}}%
}]{%
braverman2016new}
\APACinsertmetastar {%
braverman2016new}%
\begin{APACrefauthors}%
Braverman, V.%
, Feldman, D.%
\BCBL {}\ \BBA {} Lang, H.%
\end{APACrefauthors}%
\unskip\
\newblock
\APACrefYearMonthDay{2016}{}{}.
\newblock
{\BBOQ}\APACrefatitle {New Frameworks for Offline and Streaming Coreset
  Constructions} {New frameworks for offline and streaming coreset
  constructions}.{\BBCQ}
\newblock
\APACjournalVolNumPages{arXiv preprint arXiv:1612.00889}{}{}{}.
\PrintBackRefs{\CurrentBib}

\bibitem [\protect \citeauthoryear {%
B\u{a}doiu%
\ \BBA {} Clarkson%
}{%
B\u{a}doiu%
\ \BBA {} Clarkson%
}{%
{\protect \APACyear {2003}}%
}]{%
BC03}
\APACinsertmetastar {%
BC03}%
\begin{APACrefauthors}%
B\u{a}doiu, M.%
\BCBT {}\ \BBA {} Clarkson, K\BPBI L.%
\end{APACrefauthors}%
\unskip\
\newblock
\APACrefYearMonthDay{2003}{}{}.
\newblock
{\BBOQ}\APACrefatitle {Smaller core-sets for balls} {Smaller core-sets for
  balls}.{\BBCQ}
\newblock
\BIn{} \APACrefbtitle {Proceedings of the 14th SODA} {Proceedings of the 14th
  soda}\ (\BPGS\ 801 -- 802).
\PrintBackRefs{\CurrentBib}

\bibitem [\protect \citeauthoryear {%
B\u{a}doiu%
\ \BBA {} Clarkson%
}{%
B\u{a}doiu%
\ \BBA {} Clarkson%
}{%
{\protect \APACyear {2008}}%
}]{%
BC08}
\APACinsertmetastar {%
BC08}%
\begin{APACrefauthors}%
B\u{a}doiu, M.%
\BCBT {}\ \BBA {} Clarkson, K\BPBI L.%
\end{APACrefauthors}%
\unskip\
\newblock
\APACrefYearMonthDay{2008}{}{}.
\newblock
{\BBOQ}\APACrefatitle {Optimal core-sets for balls} {Optimal core-sets for
  balls}.{\BBCQ}
\newblock
\APACjournalVolNumPages{Computational Geometry}{40}{1}{14--22}.
\PrintBackRefs{\CurrentBib}

\bibitem [\protect \citeauthoryear {%
Campbell%
}{%
Campbell%
}{%
{\protect \APACyear {2018}}%
}]{%
code3}
\APACinsertmetastar {%
code3}%
\begin{APACrefauthors}%
Campbell, T.%
\end{APACrefauthors}%
\unskip\
\newblock
\APACrefYearMonthDay{2018}{}{}.
\newblock
\APACrefbtitle {Baysien coresets,
  {https://github.com/trevorcampbell/bayesian-coresets}} {Baysien coresets,
  {https://github.com/trevorcampbell/bayesian-coresets}}\
  \APACbVolEdTR{}{\BTR{}}.
\newblock
\APACaddressInstitution{}{MIT}.
\PrintBackRefs{\CurrentBib}

\bibitem [\protect \citeauthoryear {%
Charikar%
\ \BBA {} Guha%
}{%
Charikar%
\ \BBA {} Guha%
}{%
{\protect \APACyear {2005}}%
}]{%
CG05}
\APACinsertmetastar {%
CG05}%
\begin{APACrefauthors}%
Charikar, M.%
\BCBT {}\ \BBA {} Guha, S.%
\end{APACrefauthors}%
\unskip\
\newblock
\APACrefYearMonthDay{2005}{}{}.
\newblock
{\BBOQ}\APACrefatitle {Improved Combinatorial Algorithms for Facility Location
  Problems} {Improved combinatorial algorithms for facility location
  problems}.{\BBCQ}
\newblock
\APACjournalVolNumPages{SIAM Journal on Computing}{34}{4}{803 -- 824}.
\PrintBackRefs{\CurrentBib}

\bibitem [\protect \citeauthoryear {%
Charikar%
, Guha%
, Tardos%
\BCBL {}\ \BBA {} Shmoys%
}{%
Charikar%
\ \protect \BOthers {.}}{%
{\protect \APACyear {2002}}%
}]{%
CGTS02}
\APACinsertmetastar {%
CGTS02}%
\begin{APACrefauthors}%
Charikar, M.%
, Guha, S.%
, Tardos, {\'E}.%
\BCBL {}\ \BBA {} Shmoys, D\BPBI B.%
\end{APACrefauthors}%
\unskip\
\newblock
\APACrefYearMonthDay{2002}{}{}.
\newblock
{\BBOQ}\APACrefatitle {A Constant-Factor Approximation Algorithm for the
  k-Median Problem} {A constant-factor approximation algorithm for the k-median
  problem}.{\BBCQ}
\newblock
\APACjournalVolNumPages{Journal of Computer and System Sciences}{65}{1}{129 --
  149}.
\PrintBackRefs{\CurrentBib}

\bibitem [\protect \citeauthoryear {%
Chazelle%
\ \protect \BOthers {.}}{%
Chazelle%
\ \protect \BOthers {.}}{%
{\protect \APACyear {1993}}%
}]{%
Chazelle}
\APACinsertmetastar {%
Chazelle}%
\begin{APACrefauthors}%
Chazelle, B.%
, Edelsbrunner, H.%
, Grigni, M.%
, Guibas, L.%
, Sharir, M.%
\BCBL {}\ \BBA {} Welzl, E.%
\end{APACrefauthors}%
\unskip\
\newblock
\APACrefYearMonthDay{1993}{}{}.
\newblock
{\BBOQ}\APACrefatitle {Improved Bounds on Weak \&Egr;-nets for Convex Sets}
  {Improved bounds on weak \&egr;-nets for convex sets}.{\BBCQ}
\newblock
\BIn{} \APACrefbtitle {Proceedings of the Twenty-fifth Annual ACM Symposium on
  Theory of Computing {(STOC)}} {Proceedings of the twenty-fifth annual acm
  symposium on theory of computing {(STOC)}}\ (\BPGS\ 495--504).
\newblock
\APACaddressPublisher{}{ACM}.
\newblock
\begin{APACrefDOI} \doi{10.1145/167088.167222} \end{APACrefDOI}
\PrintBackRefs{\CurrentBib}

\bibitem [\protect \citeauthoryear {%
Chen%
}{%
Chen%
}{%
{\protect \APACyear {2009}}%
{\protect \APACexlab {{\protect \BCnt {1}}}}}]{%
C09}
\APACinsertmetastar {%
C09}%
\begin{APACrefauthors}%
Chen, K.%
\end{APACrefauthors}%
\unskip\
\newblock
\APACrefYearMonthDay{2009{\protect \BCnt {1}}}{}{}.
\newblock
{\BBOQ}\APACrefatitle {On Coresets for k-Median and k-Means Clustering in
  Metric and Euclidean Spaces and Their Applications} {On coresets for k-median
  and k-means clustering in metric and euclidean spaces and their
  applications}.{\BBCQ}
\newblock
\APACjournalVolNumPages{SIAM Journal on Computing}{39}{3}{923 -- 947}.
\PrintBackRefs{\CurrentBib}

\bibitem [\protect \citeauthoryear {%
Chen%
}{%
Chen%
}{%
{\protect \APACyear {2009}}%
{\protect \APACexlab {{\protect \BCnt {2}}}}}]{%
chen2009coresets}
\APACinsertmetastar {%
chen2009coresets}%
\begin{APACrefauthors}%
Chen, K.%
\end{APACrefauthors}%
\unskip\
\newblock
\APACrefYearMonthDay{2009{\protect \BCnt {2}}}{}{}.
\newblock
{\BBOQ}\APACrefatitle {On coresets for k-median and k-means clustering in
  metric and euclidean spaces and their applications} {On coresets for k-median
  and k-means clustering in metric and euclidean spaces and their
  applications}.{\BBCQ}
\newblock
\APACjournalVolNumPages{SIAM Journal on Computing}{39}{3}{923--947}.
\PrintBackRefs{\CurrentBib}

\bibitem [\protect \citeauthoryear {%
Chen%
}{%
Chen%
}{%
{\protect \APACyear {2009}}%
{\protect \APACexlab {{\protect \BCnt {3}}}}}]{%
Chen}
\APACinsertmetastar {%
Chen}%
\begin{APACrefauthors}%
Chen, K.%
\end{APACrefauthors}%
\unskip\
\newblock
\APACrefYearMonthDay{2009{\protect \BCnt {3}}}{{\APACmonth{08}}}{}.
\newblock
{\BBOQ}\APACrefatitle {On Coresets for $K$-Median and $K$-Means Clustering in
  Metric and Euclidean Spaces and Their Applications} {On coresets for
  $k$-median and $k$-means clustering in metric and euclidean spaces and their
  applications}.{\BBCQ}
\newblock
\APACjournalVolNumPages{SIAM J. Comput.}{39}{3}{923--947}.
\newblock
\begin{APACrefURL} \url{http://dx.doi.org/10.1137/070699007} \end{APACrefURL}
\newblock
\begin{APACrefDOI} \doi{10.1137/070699007} \end{APACrefDOI}
\PrintBackRefs{\CurrentBib}

\bibitem [\protect \citeauthoryear {%
Choi%
, Kim%
\BCBL {}\ \BBA {} Yu%
}{%
Choi%
\ \protect \BOthers {.}}{%
{\protect \APACyear {1997}}%
}]{%
choi1997performance}
\APACinsertmetastar {%
choi1997performance}%
\begin{APACrefauthors}%
Choi, S.%
, Kim, T.%
\BCBL {}\ \BBA {} Yu, W.%
\end{APACrefauthors}%
\unskip\
\newblock
\APACrefYearMonthDay{1997}{}{}.
\newblock
{\BBOQ}\APACrefatitle {Performance evaluation of RANSAC family} {Performance
  evaluation of ransac family}.{\BBCQ}
\newblock
\APACjournalVolNumPages{Journal of Computer Vision}{24}{3}{271--300}.
\PrintBackRefs{\CurrentBib}

\bibitem [\protect \citeauthoryear {%
Clarkson%
}{%
Clarkson%
}{%
{\protect \APACyear {2010}}%
}]{%
CL10}
\APACinsertmetastar {%
CL10}%
\begin{APACrefauthors}%
Clarkson, K\BPBI L.%
\end{APACrefauthors}%
\unskip\
\newblock
\APACrefYearMonthDay{2010}{}{}.
\newblock
{\BBOQ}\APACrefatitle {Coresets, sparse greedy approximation, and the
  Frank-Wolfe algorithm} {Coresets, sparse greedy approximation, and the
  frank-wolfe algorithm}.{\BBCQ}
\newblock
\APACjournalVolNumPages{ACM Transactions on Algorithms (TALG)}{6}{4}{63}.
\PrintBackRefs{\CurrentBib}

\bibitem [\protect \citeauthoryear {%
Clarkson%
\ \BBA {} Woodruff%
}{%
Clarkson%
\ \BBA {} Woodruff%
}{%
{\protect \APACyear {2009}}%
}]{%
CW09}
\APACinsertmetastar {%
CW09}%
\begin{APACrefauthors}%
Clarkson, K\BPBI L.%
\BCBT {}\ \BBA {} Woodruff, D\BPBI P.%
\end{APACrefauthors}%
\unskip\
\newblock
\APACrefYearMonthDay{2009}{}{}.
\newblock
{\BBOQ}\APACrefatitle {Numerical linear algebra in the streaming model}
  {Numerical linear algebra in the streaming model}.{\BBCQ}
\newblock
\BIn{} \APACrefbtitle {Proceedings of the 41st STOC} {Proceedings of the 41st
  stoc}\ (\BPGS\ 205 -- 214).
\PrintBackRefs{\CurrentBib}

\bibitem [\protect \citeauthoryear {%
Cohen%
, Elder%
, Musco%
, Musco%
\BCBL {}\ \BBA {} Persu%
}{%
Cohen%
, Elder%
\BCBL {}\ \protect \BOthers {.}}{%
{\protect \APACyear {2015}}%
}]{%
CEMMP15}
\APACinsertmetastar {%
CEMMP15}%
\begin{APACrefauthors}%
Cohen, M\BPBI B.%
, Elder, S.%
, Musco, C.%
, Musco, C.%
\BCBL {}\ \BBA {} Persu, M.%
\end{APACrefauthors}%
\unskip\
\newblock
\APACrefYearMonthDay{2015}{}{}.
\newblock
{\BBOQ}\APACrefatitle {Dimensionality Reduction for k-Means Clustering and Low
  Rank Approximation} {Dimensionality reduction for k-means clustering and low
  rank approximation}.{\BBCQ}
\newblock
\BIn{} \APACrefbtitle {Proceedings of the Forty-Seventh Annual {ACM} on
  Symposium on Theory of Computing, {STOC} 2015} {Proceedings of the
  forty-seventh annual {ACM} on symposium on theory of computing, {STOC} 2015}\
  (\BPGS\ 163--172).
\PrintBackRefs{\CurrentBib}

\bibitem [\protect \citeauthoryear {%
Cohen%
, Lee%
\BCBL {}\ \protect \BOthers {.}}{%
Cohen%
, Lee%
\BCBL {}\ \protect \BOthers {.}}{%
{\protect \APACyear {2015}}%
}]{%
Cohen15}
\APACinsertmetastar {%
Cohen15}%
\begin{APACrefauthors}%
Cohen, M\BPBI B.%
, Lee, Y\BPBI T.%
, Musco, C.%
, Musco, C.%
, Peng, R.%
\BCBL {}\ \BBA {} Sidford, A.%
\end{APACrefauthors}%
\unskip\
\newblock
\APACrefYearMonthDay{2015}{}{}.
\newblock
{\BBOQ}\APACrefatitle {Uniform Sampling for Matrix Approximation} {Uniform
  sampling for matrix approximation}.{\BBCQ}
\newblock
\BIn{} \APACrefbtitle {Proceedings of the 2015 Conference on Innovations in
  Theoretical Computer Science} {Proceedings of the 2015 conference on
  innovations in theoretical computer science}\ (\BPGS\ 181--190).
\newblock
\APACaddressPublisher{New York, NY, USA}{ACM}.
\newblock
\begin{APACrefURL} \url{http://doi.acm.org/10.1145/2688073.2688113}
  \end{APACrefURL}
\newblock
\begin{APACrefDOI} \doi{10.1145/2688073.2688113} \end{APACrefDOI}
\PrintBackRefs{\CurrentBib}

\bibitem [\protect \citeauthoryear {%
A.~Dasgupta%
, Drineas%
, Harb%
, Kumar%
\BCBL {}\ \BBA {} Mahoney%
}{%
A.~Dasgupta%
\ \protect \BOthers {.}}{%
{\protect \APACyear {2008}}%
}]{%
DasguptaDHKM08}
\APACinsertmetastar {%
DasguptaDHKM08}%
\begin{APACrefauthors}%
Dasgupta, A.%
, Drineas, P.%
, Harb, B.%
, Kumar, R.%
\BCBL {}\ \BBA {} Mahoney, M\BPBI W.%
\end{APACrefauthors}%
\unskip\
\newblock
\APACrefYearMonthDay{2008}{}{}.
\newblock
{\BBOQ}\APACrefatitle {Sampling algorithms and coresets for
  {$\ell_p$}-regression} {Sampling algorithms and coresets for
  {$\ell_p$}-regression}.{\BBCQ}
\newblock
\BIn{} \APACrefbtitle {Proc. 19th Annu. {ACM}-{SIAM} Symp. on Discrete
  Algorithms ({SODA})} {Proc. 19th annu. {ACM}-{SIAM} symp. on discrete
  algorithms ({SODA})}\ (\BPGS\ 932--941).
\newblock
\begin{APACrefURL} \url{http://doi.acm.org/10.1145/1347082.1347184}
  \end{APACrefURL}
\PrintBackRefs{\CurrentBib}

\bibitem [\protect \citeauthoryear {%
S.~Dasgupta%
\ \BBA {} Schulman%
}{%
S.~Dasgupta%
\ \BBA {} Schulman%
}{%
{\protect \APACyear {2000}}%
}]{%
dasgupta2000two}
\APACinsertmetastar {%
dasgupta2000two}%
\begin{APACrefauthors}%
Dasgupta, S.%
\BCBT {}\ \BBA {} Schulman, L\BPBI J.%
\end{APACrefauthors}%
\unskip\
\newblock
\APACrefYearMonthDay{2000}{}{}.
\newblock
{\BBOQ}\APACrefatitle {A two-round variant of em for gaussian mixtures} {A
  two-round variant of em for gaussian mixtures}.{\BBCQ}
\newblock
\BIn{} \APACrefbtitle {Proceedings of the Sixteenth conference on Uncertainty
  in artificial intelligence} {Proceedings of the sixteenth conference on
  uncertainty in artificial intelligence}\ (\BPGS\ 152--159).
\PrintBackRefs{\CurrentBib}

\bibitem [\protect \citeauthoryear {%
Deshpande%
, Rademacher%
, Vempala%
\BCBL {}\ \BBA {} Wang%
}{%
Deshpande%
\ \protect \BOthers {.}}{%
{\protect \APACyear {2006}}%
}]{%
santosh}
\APACinsertmetastar {%
santosh}%
\begin{APACrefauthors}%
Deshpande, A.%
, Rademacher, L.%
, Vempala, S.%
\BCBL {}\ \BBA {} Wang, G.%
\end{APACrefauthors}%
\unskip\
\newblock
\APACrefYearMonthDay{2006}{}{}.
\newblock
{\BBOQ}\APACrefatitle {Matrix Approximation and Projective Clustering via
  Volume Sampling} {Matrix approximation and projective clustering via volume
  sampling}.{\BBCQ}
\newblock
\BIn{} \APACrefbtitle {Proc. 17th Annu. {ACM}-{SIAM} Symp. on Discrete
  Algorithms ({SODA})} {Proc. 17th annu. {ACM}-{SIAM} symp. on discrete
  algorithms ({SODA})}\ (\BPGS\ 1117--1126).
\PrintBackRefs{\CurrentBib}

\bibitem [\protect \citeauthoryear {%
Drineas%
, Mahoney%
\BCBL {}\ \BBA {} Muthukrishnan%
}{%
Drineas%
\ \protect \BOthers {.}}{%
{\protect \APACyear {2006}}%
}]{%
DrineasMM06}
\APACinsertmetastar {%
DrineasMM06}%
\begin{APACrefauthors}%
Drineas, P.%
, Mahoney, M\BPBI W.%
\BCBL {}\ \BBA {} Muthukrishnan, S.%
\end{APACrefauthors}%
\unskip\
\newblock
\APACrefYearMonthDay{2006}{}{}.
\newblock
{\BBOQ}\APACrefatitle {Sampling algorithms for $l_2$ regression and
  applications} {Sampling algorithms for $l_2$ regression and
  applications}.{\BBCQ}
\newblock
\BIn{} \APACrefbtitle {SODA.} {Soda.}
\PrintBackRefs{\CurrentBib}

\bibitem [\protect \citeauthoryear {%
Edwards%
\ \BBA {} Varadarajan%
}{%
Edwards%
\ \BBA {} Varadarajan%
}{%
{\protect \APACyear {2005}}%
}]{%
edwards2005no}
\APACinsertmetastar {%
edwards2005no}%
\begin{APACrefauthors}%
Edwards, M.%
\BCBT {}\ \BBA {} Varadarajan, K.%
\end{APACrefauthors}%
\unskip\
\newblock
\APACrefYearMonthDay{2005}{}{}.
\newblock
{\BBOQ}\APACrefatitle {No coreset, no cry: II} {No coreset, no cry: Ii}.{\BBCQ}
\newblock
\BIn{} \APACrefbtitle {International Conference on Foundations of Software
  Technology and Theoretical Computer Science} {International conference on
  foundations of software technology and theoretical computer science}\ (\BPGS\
  107--115).
\PrintBackRefs{\CurrentBib}

\bibitem [\protect \citeauthoryear {%
Effros%
\ \BBA {} Schulman%
}{%
Effros%
\ \BBA {} Schulman%
}{%
{\protect \APACyear {2004}}%
}]{%
ES04}
\APACinsertmetastar {%
ES04}%
\begin{APACrefauthors}%
Effros, M.%
\BCBT {}\ \BBA {} Schulman, L\BPBI J.%
\end{APACrefauthors}%
\unskip\
\newblock
\APACrefYearMonthDay{2004}{}{}.
\newblock
{\BBOQ}\APACrefatitle {Deterministic clustering with data nets} {Deterministic
  clustering with data nets}.{\BBCQ}
\newblock
\APACjournalVolNumPages{Electronic Colloquium on Computational Complexity
  (ECCC)}{}{050}{}.
\PrintBackRefs{\CurrentBib}

\bibitem [\protect \citeauthoryear {%
Epstein%
\ \BBA {} Feldman%
}{%
Epstein%
\ \BBA {} Feldman%
}{%
{\protect \APACyear {2018}}%
}]{%
epstein2018quadcopter}
\APACinsertmetastar {%
epstein2018quadcopter}%
\begin{APACrefauthors}%
Epstein, D.%
\BCBT {}\ \BBA {} Feldman, D.%
\end{APACrefauthors}%
\unskip\
\newblock
\APACrefYearMonthDay{2018}{}{}.
\newblock
{\BBOQ}\APACrefatitle {Quadcopter Tracks Quadcopter via Real-Time Shape
  Fitting} {Quadcopter tracks quadcopter via real-time shape fitting}.{\BBCQ}
\newblock
\APACjournalVolNumPages{IEEE Robotics and Automation Letters}{3}{1}{544--550}.
\PrintBackRefs{\CurrentBib}

\bibitem [\protect \citeauthoryear {%
Feigin%
, Feldman%
\BCBL {}\ \BBA {} Sochen%
}{%
Feigin%
\ \protect \BOthers {.}}{%
{\protect \APACyear {2011}}%
}]{%
ffs11}
\APACinsertmetastar {%
ffs11}%
\begin{APACrefauthors}%
Feigin, M.%
, Feldman, D.%
\BCBL {}\ \BBA {} Sochen, N.%
\end{APACrefauthors}%
\unskip\
\newblock
\APACrefYearMonthDay{2011}{}{}.
\newblock
{\BBOQ}\APACrefatitle {From High Definition Image to Low Space Optimization}
  {From high definition image to low space optimization}.{\BBCQ}
\newblock
\BIn{} \APACrefbtitle {Proc. 3rd Inter. Conf. on Scale Space and Variational
  Methods in Computer Vision ({SSVM} 2011).} {Proc. 3rd inter. conf. on scale
  space and variational methods in computer vision ({SSVM} 2011).}
\PrintBackRefs{\CurrentBib}

\bibitem [\protect \citeauthoryear {%
Feldman%
}{%
Feldman%
}{%
{\protect \APACyear {2018}}%
}]{%
code2}
\APACinsertmetastar {%
code2}%
\begin{APACrefauthors}%
Feldman, D.%
\end{APACrefauthors}%
\unskip\
\newblock
\APACrefYearMonthDay{2018}{}{}.
\newblock
\APACrefbtitle {Code for recent coresets papers,
  {https://sites.hevra.haifa.ac.il/rbd/}} {Code for recent coresets papers,
  {https://sites.hevra.haifa.ac.il/rbd/}}\ \APACbVolEdTR{}{\BTR{}}.
\newblock
\APACaddressInstitution{}{Robotics \& Big Data Lab, University of Haifa}.
\PrintBackRefs{\CurrentBib}

\bibitem [\protect \citeauthoryear {%
Feldman%
, Fiat%
, Kaplan%
\BCBL {}\ \BBA {} Nissim%
}{%
Feldman%
\ \protect \BOthers {.}}{%
{\protect \APACyear {2009}}%
}]{%
feldman2009private}
\APACinsertmetastar {%
feldman2009private}%
\begin{APACrefauthors}%
Feldman, D.%
, Fiat, A.%
, Kaplan, H.%
\BCBL {}\ \BBA {} Nissim, K.%
\end{APACrefauthors}%
\unskip\
\newblock
\APACrefYearMonthDay{2009}{}{}.
\newblock
{\BBOQ}\APACrefatitle {Private coresets} {Private coresets}.{\BBCQ}
\newblock
\BIn{} \APACrefbtitle {Proceedings of the forty-first annual ACM symposium on
  Theory of computing} {Proceedings of the forty-first annual acm symposium on
  theory of computing}\ (\BPGS\ 361--370).
\PrintBackRefs{\CurrentBib}

\bibitem [\protect \citeauthoryear {%
Feldman%
, Fiat%
, Segev%
\BCBL {}\ \BBA {} Sharir%
}{%
Feldman%
, Fiat%
\BCBL {}\ \protect \BOthers {.}}{%
{\protect \APACyear {2007}}%
}]{%
FFSS07}
\APACinsertmetastar {%
FFSS07}%
\begin{APACrefauthors}%
Feldman, D.%
, Fiat, A.%
, Segev, D.%
\BCBL {}\ \BBA {} Sharir, M.%
\end{APACrefauthors}%
\unskip\
\newblock
\APACrefYearMonthDay{2007}{}{}.
\newblock
{\BBOQ}\APACrefatitle {Bi-criteria linear-time approximations for generalized
  k-mean/median/center} {Bi-criteria linear-time approximations for generalized
  k-mean/median/center}.{\BBCQ}
\newblock
\BIn{} \APACrefbtitle {Proc. 23rd {ACM} Symp. on Computational Geometry
  ({SOCG})} {Proc. 23rd {ACM} symp. on computational geometry ({SOCG})}\
  (\BPGS\ 19--26).
\PrintBackRefs{\CurrentBib}

\bibitem [\protect \citeauthoryear {%
Feldman%
, Fiat%
\BCBL {}\ \BBA {} Sharir%
}{%
Feldman%
\ \protect \BOthers {.}}{%
{\protect \APACyear {2006}}%
}]{%
FFS06}
\APACinsertmetastar {%
FFS06}%
\begin{APACrefauthors}%
Feldman, D.%
, Fiat, A.%
\BCBL {}\ \BBA {} Sharir, M.%
\end{APACrefauthors}%
\unskip\
\newblock
\APACrefYearMonthDay{2006}{}{}.
\newblock
{\BBOQ}\APACrefatitle {Coresets for Weighted Facilities and Their Applications}
  {Coresets for weighted facilities and their applications}.{\BBCQ}
\newblock
\BIn{} \APACrefbtitle {FOCS} {Focs}\ (\BPGS\ 315--324).
\PrintBackRefs{\CurrentBib}

\bibitem [\protect \citeauthoryear {%
Feldman%
, Krause%
\BCBL {}\ \BBA {} Faulkner%
}{%
Feldman%
\ \protect \BOthers {.}}{%
{\protect \APACyear {2011}}%
}]{%
NIPS}
\APACinsertmetastar {%
NIPS}%
\begin{APACrefauthors}%
Feldman, D.%
, Krause, A.%
\BCBL {}\ \BBA {} Faulkner, M.%
\end{APACrefauthors}%
\unskip\
\newblock
\APACrefYearMonthDay{2011}{}{}.
\newblock
{\BBOQ}\APACrefatitle {Scalable Training of Mixture Models via Coresets}
  {Scalable training of mixture models via coresets}.{\BBCQ}
\newblock
\BIn{} \APACrefbtitle {Proc. 25th Conference on Neural Information Processing
  Systems (NIPS).} {Proc. 25th conference on neural information processing
  systems (nips).}
\PrintBackRefs{\CurrentBib}

\bibitem [\protect \citeauthoryear {%
Feldman%
\ \BBA {} Langberg%
}{%
Feldman%
\ \BBA {} Langberg%
}{%
{\protect \APACyear {2011}}%
}]{%
FL11}
\APACinsertmetastar {%
FL11}%
\begin{APACrefauthors}%
Feldman, D.%
\BCBT {}\ \BBA {} Langberg, M.%
\end{APACrefauthors}%
\unskip\
\newblock
\APACrefYearMonthDay{2011}{}{}.
\newblock
{\BBOQ}\APACrefatitle {A Unified Framework for Approximating and Clustering
  Data.} {A unified framework for approximating and clustering data.}{\BBCQ}
\newblock
\BIn{} \APACrefbtitle {Proc. 34th Annu. {ACM} Symp. on Theory of Computing
  ({STOC}).} {Proc. 34th annu. {ACM} symp. on theory of computing ({STOC}).}
\newblock
\APACrefnote{See http://arxiv.org/abs/1106.1379 for fuller version.}
\PrintBackRefs{\CurrentBib}

\bibitem [\protect \citeauthoryear {%
Feldman%
, Monemizadeh%
\BCBL {}\ \BBA {} Sohler%
}{%
Feldman%
, Monemizadeh%
\BCBL {}\ \BBA {} Sohler%
}{%
{\protect \APACyear {2007}}%
}]{%
FMS07}
\APACinsertmetastar {%
FMS07}%
\begin{APACrefauthors}%
Feldman, D.%
, Monemizadeh, M.%
\BCBL {}\ \BBA {} Sohler, C.%
\end{APACrefauthors}%
\unskip\
\newblock
\APACrefYearMonthDay{2007}{}{}.
\newblock
{\BBOQ}\APACrefatitle {A PTAS for k-means clustering based on weak coresets} {A
  ptas for k-means clustering based on weak coresets}.{\BBCQ}
\newblock
\BIn{} \APACrefbtitle {\Proc 23rd \SoCG} {\proc 23rd \socg}\ (\BPGS\ 11--18).
\PrintBackRefs{\CurrentBib}

\bibitem [\protect \citeauthoryear {%
Feldman%
, Monemizadeh%
, Sohler%
\BCBL {}\ \BBA {} Woodruff%
}{%
Feldman%
\ \protect \BOthers {.}}{%
{\protect \APACyear {2010}}%
}]{%
feldman2010coresets}
\APACinsertmetastar {%
feldman2010coresets}%
\begin{APACrefauthors}%
Feldman, D.%
, Monemizadeh, M.%
, Sohler, C.%
\BCBL {}\ \BBA {} Woodruff, D\BPBI P.%
\end{APACrefauthors}%
\unskip\
\newblock
\APACrefYearMonthDay{2010}{}{}.
\newblock
{\BBOQ}\APACrefatitle {Coresets and sketches for high dimensional subspace
  approximation problems} {Coresets and sketches for high dimensional subspace
  approximation problems}.{\BBCQ}
\newblock
\BIn{} \APACrefbtitle {Proceedings of the twenty-first annual ACM-SIAM
  symposium on Discrete Algorithms} {Proceedings of the twenty-first annual
  acm-siam symposium on discrete algorithms}\ (\BPGS\ 630--649).
\PrintBackRefs{\CurrentBib}

\bibitem [\protect \citeauthoryear {%
Feldman%
, Ozer%
\BCBL {}\ \BBA {} Rus%
}{%
Feldman%
, Ozer%
\BCBL {}\ \BBA {} Rus%
}{%
{\protect \APACyear {2017}}%
}]{%
FeldmanOR17}
\APACinsertmetastar {%
FeldmanOR17}%
\begin{APACrefauthors}%
Feldman, D.%
, Ozer, S.%
\BCBL {}\ \BBA {} Rus, D.%
\end{APACrefauthors}%
\unskip\
\newblock
\APACrefYearMonthDay{2017}{}{}.
\newblock
{\BBOQ}\APACrefatitle {Coresets for Vector Summarization with Applications to
  Network Graphs} {Coresets for vector summarization with applications to
  network graphs}.{\BBCQ}
\newblock
\BIn{} \APACrefbtitle {Proceedings of the 34th International Conference on
  Machine Learning, {ICML} 2017, Sydney, NSW, Australia, 6-11 August 2017}
  {Proceedings of the 34th international conference on machine learning, {ICML}
  2017, sydney, nsw, australia, 6-11 august 2017}\ (\BPGS\ 1117--1125).
\newblock
\begin{APACrefURL} \url{http://proceedings.mlr.press/v70/feldman17a.html}
  \end{APACrefURL}
\PrintBackRefs{\CurrentBib}

\bibitem [\protect \citeauthoryear {%
Feldman%
, Schmidt%
\BCBL {}\ \BBA {} Sohler%
}{%
Feldman%
, Schmidt%
\BCBL {}\ \BBA {} Sohler%
}{%
{\protect \APACyear {2013}}%
{\protect \APACexlab {{\protect \BCnt {1}}}}}]{%
FSS13}
\APACinsertmetastar {%
FSS13}%
\begin{APACrefauthors}%
Feldman, D.%
, Schmidt, M.%
\BCBL {}\ \BBA {} Sohler, C.%
\end{APACrefauthors}%
\unskip\
\newblock
\APACrefYearMonthDay{2013{\protect \BCnt {1}}}{}{}.
\newblock
{\BBOQ}\APACrefatitle {Turning big data into tiny data: Constant-size coresets
  for k-means, pca and projective clustering} {Turning big data into tiny data:
  Constant-size coresets for k-means, pca and projective clustering}.{\BBCQ}
\newblock
\BIn{} \APACrefbtitle {Proceedings of the Twenty-Fourth Annual ACM-SIAM
  Symposium on Discrete Algorithms} {Proceedings of the twenty-fourth annual
  acm-siam symposium on discrete algorithms}\ (\BPGS\ 1434--1453).
\PrintBackRefs{\CurrentBib}

\bibitem [\protect \citeauthoryear {%
Feldman%
, Schmidt%
\BCBL {}\ \BBA {} Sohler%
}{%
Feldman%
, Schmidt%
\BCBL {}\ \BBA {} Sohler%
}{%
{\protect \APACyear {2013}}%
{\protect \APACexlab {{\protect \BCnt {2}}}}}]{%
feldman2013turning}
\APACinsertmetastar {%
feldman2013turning}%
\begin{APACrefauthors}%
Feldman, D.%
, Schmidt, M.%
\BCBL {}\ \BBA {} Sohler, C.%
\end{APACrefauthors}%
\unskip\
\newblock
\APACrefYearMonthDay{2013{\protect \BCnt {2}}}{}{}.
\newblock
{\BBOQ}\APACrefatitle {Turning big data into tiny data: Constant-size coresets
  for k-means, pca and projective clustering} {Turning big data into tiny data:
  Constant-size coresets for k-means, pca and projective clustering}.{\BBCQ}
\newblock
\BIn{} \APACrefbtitle {Proceedings of the Twenty-Fourth Annual ACM-SIAM
  Symposium on Discrete Algorithms} {Proceedings of the twenty-fourth annual
  acm-siam symposium on discrete algorithms}\ (\BPGS\ 1434--1453).
\PrintBackRefs{\CurrentBib}

\bibitem [\protect \citeauthoryear {%
Feldman%
\ \BBA {} Schulman%
}{%
Feldman%
\ \BBA {} Schulman%
}{%
{\protect \APACyear {2012}}%
{\protect \APACexlab {{\protect \BCnt {1}}}}}]{%
feldman2012data}
\APACinsertmetastar {%
feldman2012data}%
\begin{APACrefauthors}%
Feldman, D.%
\BCBT {}\ \BBA {} Schulman, L\BPBI J.%
\end{APACrefauthors}%
\unskip\
\newblock
\APACrefYearMonthDay{2012{\protect \BCnt {1}}}{}{}.
\newblock
{\BBOQ}\APACrefatitle {Data reduction for weighted and outlier-resistant
  clustering} {Data reduction for weighted and outlier-resistant
  clustering}.{\BBCQ}
\newblock
\BIn{} \APACrefbtitle {Proceedings of the twenty-third annual ACM-SIAM
  symposium on Discrete Algorithms} {Proceedings of the twenty-third annual
  acm-siam symposium on discrete algorithms}\ (\BPGS\ 1343--1354).
\PrintBackRefs{\CurrentBib}

\bibitem [\protect \citeauthoryear {%
Feldman%
\ \BBA {} Schulman%
}{%
Feldman%
\ \BBA {} Schulman%
}{%
{\protect \APACyear {2012}}%
{\protect \APACexlab {{\protect \BCnt {2}}}}}]{%
schulman}
\APACinsertmetastar {%
schulman}%
\begin{APACrefauthors}%
Feldman, D.%
\BCBT {}\ \BBA {} Schulman, L\BPBI J.%
\end{APACrefauthors}%
\unskip\
\newblock
\APACrefYearMonthDay{2012{\protect \BCnt {2}}}{}{}.
\newblock
{\BBOQ}\APACrefatitle {Data reduction for weighted and outlier-resistant
  clustering} {Data reduction for weighted and outlier-resistant
  clustering}.{\BBCQ}
\newblock
\BIn{} \APACrefbtitle {Proc. of the 23rd annual ACM-SIAM symp. on Discrete
  Algorithms (SODA)} {Proc. of the 23rd annual acm-siam symp. on discrete
  algorithms (soda)}\ (\BPGS\ 1343--1354).
\PrintBackRefs{\CurrentBib}

\bibitem [\protect \citeauthoryear {%
Feldman%
, Sugaya%
\BCBL {}\ \BBA {} Rus%
}{%
Feldman%
, Sugaya%
\BCBL {}\ \BBA {} Rus%
}{%
{\protect \APACyear {2012}}%
}]{%
feldman2012effective}
\APACinsertmetastar {%
feldman2012effective}%
\begin{APACrefauthors}%
Feldman, D.%
, Sugaya, A.%
\BCBL {}\ \BBA {} Rus, D.%
\end{APACrefauthors}%
\unskip\
\newblock
\APACrefYearMonthDay{2012}{}{}.
\newblock
{\BBOQ}\APACrefatitle {An effective coreset compression algorithm for large
  scale sensor networks} {An effective coreset compression algorithm for large
  scale sensor networks}.{\BBCQ}
\newblock
\BIn{} \APACrefbtitle {Information Processing in Sensor Networks (IPSN), 2012
  ACM/IEEE 11th International Conference on} {Information processing in sensor
  networks (ipsn), 2012 acm/ieee 11th international conference on}\ (\BPGS\
  257--268).
\PrintBackRefs{\CurrentBib}

\bibitem [\protect \citeauthoryear {%
Feldman%
, Sugaya%
, Sung%
\BCBL {}\ \BBA {} Rus%
}{%
Feldman%
, Sugaya%
\BCBL {}\ \protect \BOthers {.}}{%
{\protect \APACyear {2013}}%
}]{%
feldman2013idiary}
\APACinsertmetastar {%
feldman2013idiary}%
\begin{APACrefauthors}%
Feldman, D.%
, Sugaya, A.%
, Sung, C.%
\BCBL {}\ \BBA {} Rus, D.%
\end{APACrefauthors}%
\unskip\
\newblock
\APACrefYearMonthDay{2013}{}{}.
\newblock
{\BBOQ}\APACrefatitle {iDiary: from GPS signals to a text-searchable diary}
  {idiary: from gps signals to a text-searchable diary}.{\BBCQ}
\newblock
\BIn{} \APACrefbtitle {Proceedings of the 11th ACM Conference on Embedded
  Networked Sensor Systems} {Proceedings of the 11th acm conference on embedded
  networked sensor systems}\ (\BPG~6).
\PrintBackRefs{\CurrentBib}

\bibitem [\protect \citeauthoryear {%
Feldman%
, Sung%
\BCBL {}\ \BBA {} Rus%
}{%
Feldman%
, Sung%
\BCBL {}\ \BBA {} Rus%
}{%
{\protect \APACyear {2012}}%
}]{%
feldman2012single}
\APACinsertmetastar {%
feldman2012single}%
\begin{APACrefauthors}%
Feldman, D.%
, Sung, C.%
\BCBL {}\ \BBA {} Rus, D.%
\end{APACrefauthors}%
\unskip\
\newblock
\APACrefYearMonthDay{2012}{}{}.
\newblock
{\BBOQ}\APACrefatitle {The single pixel GPS: learning big data signals from
  tiny coresets} {The single pixel gps: learning big data signals from tiny
  coresets}.{\BBCQ}
\newblock
\BIn{} \APACrefbtitle {Proceedings of the 20th International Conference on
  Advances in Geographic Information Systems} {Proceedings of the 20th
  international conference on advances in geographic information systems}\
  (\BPGS\ 23--32).
\PrintBackRefs{\CurrentBib}

\bibitem [\protect \citeauthoryear {%
Feldman%
\ \BBA {} Tassa%
}{%
Feldman%
\ \BBA {} Tassa%
}{%
{\protect \APACyear {2015}}%
}]{%
FT15}
\APACinsertmetastar {%
FT15}%
\begin{APACrefauthors}%
Feldman, D.%
\BCBT {}\ \BBA {} Tassa, T.%
\end{APACrefauthors}%
\unskip\
\newblock
\APACrefYearMonthDay{2015}{}{}.
\newblock
{\BBOQ}\APACrefatitle {More constraints, smaller coresets: constrained matrix
  approximation of sparse big data} {More constraints, smaller coresets:
  constrained matrix approximation of sparse big data}.{\BBCQ}
\newblock
\BIn{} \APACrefbtitle {Proceedings of the 21th ACM SIGKDD International
  Conference on Knowledge Discovery and Data Mining (KDD'15)} {Proceedings of
  the 21th acm sigkdd international conference on knowledge discovery and data
  mining (kdd'15)}\ (\BPGS\ 249--258).
\PrintBackRefs{\CurrentBib}

\bibitem [\protect \citeauthoryear {%
Feldman%
, Volkov%
\BCBL {}\ \BBA {} Rus%
}{%
Feldman%
\ \protect \BOthers {.}}{%
{\protect \APACyear {2016}}%
}]{%
feldmanmik}
\APACinsertmetastar {%
feldmanmik}%
\begin{APACrefauthors}%
Feldman, D.%
, Volkov, M.%
\BCBL {}\ \BBA {} Rus, D.%
\end{APACrefauthors}%
\unskip\
\newblock
\APACrefYearMonthDay{2016}{}{}.
\newblock
{\BBOQ}\APACrefatitle {Dimensionality Reduction of Massive Sparse Datasets
  Using Coresets} {Dimensionality reduction of massive sparse datasets using
  coresets}.{\BBCQ}
\newblock
\BIn{} \APACrefbtitle {Advances in neural information processing systems
  (NIPS).} {Advances in neural information processing systems (nips).}
\PrintBackRefs{\CurrentBib}

\bibitem [\protect \citeauthoryear {%
Feldman%
, Xiang%
, Zhu%
\BCBL {}\ \BBA {} Rus%
}{%
Feldman%
, Xiang%
\BCBL {}\ \protect \BOthers {.}}{%
{\protect \APACyear {2017}}%
}]{%
feldman2017coresets}
\APACinsertmetastar {%
feldman2017coresets}%
\begin{APACrefauthors}%
Feldman, D.%
, Xiang, C.%
, Zhu, R.%
\BCBL {}\ \BBA {} Rus, D.%
\end{APACrefauthors}%
\unskip\
\newblock
\APACrefYearMonthDay{2017}{}{}.
\newblock
{\BBOQ}\APACrefatitle {Coresets for differentially private k-means clustering
  and applications to privacy in mobile sensor networks} {Coresets for
  differentially private k-means clustering and applications to privacy in
  mobile sensor networks}.{\BBCQ}
\newblock
\BIn{} \APACrefbtitle {Information Processing in Sensor Networks (IPSN), 2017
  16th ACM/IEEE International Conference on} {Information processing in sensor
  networks (ipsn), 2017 16th acm/ieee international conference on}\ (\BPGS\
  3--16).
\PrintBackRefs{\CurrentBib}

\bibitem [\protect \citeauthoryear {%
Foster%
}{%
Foster%
}{%
{\protect \APACyear {1995}}%
}]{%
foster1995designing}
\APACinsertmetastar {%
foster1995designing}%
\begin{APACrefauthors}%
Foster, I.%
\end{APACrefauthors}%
\unskip\
\newblock
\APACrefYear{1995}.
\newblock
\APACrefbtitle {Designing and building parallel programs} {Designing and
  building parallel programs}.
\newblock
\APACaddressPublisher{}{Addison Wesley Publishing Company}.
\PrintBackRefs{\CurrentBib}

\bibitem [\protect \citeauthoryear {%
Funke%
\ \BBA {} Laue%
}{%
Funke%
\ \BBA {} Laue%
}{%
{\protect \APACyear {2007}}%
}]{%
funke2007bounded}
\APACinsertmetastar {%
funke2007bounded}%
\begin{APACrefauthors}%
Funke, S.%
\BCBT {}\ \BBA {} Laue, S.%
\end{APACrefauthors}%
\unskip\
\newblock
\APACrefYearMonthDay{2007}{}{}.
\newblock
{\BBOQ}\APACrefatitle {Bounded-hop energy-efficient broadcast in
  low-dimensional metrics via coresets} {Bounded-hop energy-efficient broadcast
  in low-dimensional metrics via coresets}.{\BBCQ}
\newblock
\BIn{} \APACrefbtitle {Annual Symposium on Theoretical Aspects of Computer
  Science} {Annual symposium on theoretical aspects of computer science}\
  (\BPGS\ 272--283).
\PrintBackRefs{\CurrentBib}

\bibitem [\protect \citeauthoryear {%
Har{-}Peled%
}{%
Har{-}Peled%
}{%
{\protect \APACyear {2004}}%
}]{%
H04}
\APACinsertmetastar {%
H04}%
\begin{APACrefauthors}%
Har{-}Peled, S.%
\end{APACrefauthors}%
\unskip\
\newblock
\APACrefYearMonthDay{2004}{}{}.
\newblock
{\BBOQ}\APACrefatitle {No Coreset, No Cry} {No coreset, no cry}.{\BBCQ}
\newblock
\BIn{} \APACrefbtitle {\Proc 24th \FSTTCS} {\proc 24th \fsttcs}\ (\BPGS\
  324--335).
\PrintBackRefs{\CurrentBib}

\bibitem [\protect \citeauthoryear {%
Har{-}Peled%
}{%
Har{-}Peled%
}{%
{\protect \APACyear {2006}}%
}]{%
HP06}
\APACinsertmetastar {%
HP06}%
\begin{APACrefauthors}%
Har{-}Peled, S.%
\end{APACrefauthors}%
\unskip\
\newblock
\APACrefYearMonthDay{2006}{}{}.
\newblock
{\BBOQ}\APACrefatitle {Coresets for Discrete Integration and Clustering}
  {Coresets for discrete integration and clustering}.{\BBCQ}
\newblock
\BIn{} \APACrefbtitle {26th FSTTCS} {26th fsttcs}\ (\BPGS\ 33 -- 44).
\PrintBackRefs{\CurrentBib}

\bibitem [\protect \citeauthoryear {%
Har-Peled%
\ \BBA {} Kushal%
}{%
Har-Peled%
\ \BBA {} Kushal%
}{%
{\protect \APACyear {2005}}%
}]{%
HK05}
\APACinsertmetastar {%
HK05}%
\begin{APACrefauthors}%
Har-Peled, S.%
\BCBT {}\ \BBA {} Kushal, A.%
\end{APACrefauthors}%
\unskip\
\newblock
\APACrefYearMonthDay{2005}{}{}.
\newblock
{\BBOQ}\APACrefatitle {Smaller coresets for $k$-median and $k$-means
  clustering} {Smaller coresets for $k$-median and $k$-means
  clustering}.{\BBCQ}
\newblock
\BIn{} \APACrefbtitle {Proceedings of the 25th SODA} {Proceedings of the 25th
  soda}\ (\BPGS\ 126--134).
\PrintBackRefs{\CurrentBib}

\bibitem [\protect \citeauthoryear {%
Har-Peled%
\ \BBA {} Kushal%
}{%
Har-Peled%
\ \BBA {} Kushal%
}{%
{\protect \APACyear {2007}}%
}]{%
sarielb}
\APACinsertmetastar {%
sarielb}%
\begin{APACrefauthors}%
Har-Peled, S.%
\BCBT {}\ \BBA {} Kushal, A.%
\end{APACrefauthors}%
\unskip\
\newblock
\APACrefYearMonthDay{2007}{}{}.
\newblock
{\BBOQ}\APACrefatitle {Smaller coresets for $k$-median and $k$-means
  clustering} {Smaller coresets for $k$-median and $k$-means
  clustering}.{\BBCQ}
\newblock
\APACjournalVolNumPages{Discrete Comput. Geom.}{37}{1}{3--19}.
\newblock
\begin{APACrefURL} \url{http://dx.doi.org/10.1007/s00454-006-1271-x}
  \end{APACrefURL}
\PrintBackRefs{\CurrentBib}

\bibitem [\protect \citeauthoryear {%
Har-Peled%
\ \BBA {} Mazumdar%
}{%
Har-Peled%
\ \BBA {} Mazumdar%
}{%
{\protect \APACyear {2004}}%
{\protect \APACexlab {{\protect \BCnt {1}}}}}]{%
HM04}
\APACinsertmetastar {%
HM04}%
\begin{APACrefauthors}%
Har-Peled, S.%
\BCBT {}\ \BBA {} Mazumdar, S.%
\end{APACrefauthors}%
\unskip\
\newblock
\APACrefYearMonthDay{2004{\protect \BCnt {1}}}{}{}.
\newblock
{\BBOQ}\APACrefatitle {Coresets for $k$-means and $k$-median clustering and
  their applications} {Coresets for $k$-means and $k$-median clustering and
  their applications}.{\BBCQ}
\newblock
\BIn{} \APACrefbtitle {\Proc 36th \STOC} {\proc 36th \stoc}\ (\BPGS\ 291--300).
\PrintBackRefs{\CurrentBib}

\bibitem [\protect \citeauthoryear {%
Har-Peled%
\ \BBA {} Mazumdar%
}{%
Har-Peled%
\ \BBA {} Mazumdar%
}{%
{\protect \APACyear {2004}}%
{\protect \APACexlab {{\protect \BCnt {2}}}}}]{%
sariela}
\APACinsertmetastar {%
sariela}%
\begin{APACrefauthors}%
Har-Peled, S.%
\BCBT {}\ \BBA {} Mazumdar, S.%
\end{APACrefauthors}%
\unskip\
\newblock
\APACrefYearMonthDay{2004{\protect \BCnt {2}}}{}{}.
\newblock
{\BBOQ}\APACrefatitle {On coresets for k-means and k-median clustering} {On
  coresets for k-means and k-median clustering}.{\BBCQ}
\newblock
\BIn{} \APACrefbtitle {Proc. 36th Ann. {ACM} Symp. on Theory of Computing
  ({STOC})} {Proc. 36th ann. {ACM} symp. on theory of computing ({STOC})}\
  (\BPGS\ 291--300).
\PrintBackRefs{\CurrentBib}

\bibitem [\protect \citeauthoryear {%
Har-Peled%
\ \BBA {} Varadarajan%
}{%
Har-Peled%
\ \BBA {} Varadarajan%
}{%
{\protect \APACyear {2002}}%
}]{%
HV02}
\APACinsertmetastar {%
HV02}%
\begin{APACrefauthors}%
Har-Peled, S.%
\BCBT {}\ \BBA {} Varadarajan, K\BPBI R.%
\end{APACrefauthors}%
\unskip\
\newblock
\APACrefYearMonthDay{2002}{}{}.
\newblock
{\BBOQ}\APACrefatitle {Projective clustering in high dimensions using
  coresets.} {Projective clustering in high dimensions using coresets.}{\BBCQ}
\newblock
\BIn{} \APACrefbtitle {Proc. 18th {ACM} Symp. on Computational Geometry (SoCG)}
  {Proc. 18th {ACM} symp. on computational geometry (socg)}\ (\BPGS\ 312--318).
\PrintBackRefs{\CurrentBib}

\bibitem [\protect \citeauthoryear {%
Haussler%
}{%
Haussler%
}{%
{\protect \APACyear {1990}}%
}]{%
H90}
\APACinsertmetastar {%
H90}%
\begin{APACrefauthors}%
Haussler, D.%
\end{APACrefauthors}%
\unskip\
\newblock
\APACrefYearMonthDay{1990}{}{}.
\newblock
{\BBOQ}\APACrefatitle {Decision Theoretic Generalizations of the PAC Learning
  Model} {Decision theoretic generalizations of the pac learning model}.{\BBCQ}
\newblock
\BIn{} \APACrefbtitle {\Proc 1st \ALT} {\proc 1st \alt}\ (\BPGS\ 21 -- 41).
\PrintBackRefs{\CurrentBib}

\bibitem [\protect \citeauthoryear {%
Haussler%
\ \BBA {} Welzl%
}{%
Haussler%
\ \BBA {} Welzl%
}{%
{\protect \APACyear {1986}}%
}]{%
HauWel86}
\APACinsertmetastar {%
HauWel86}%
\begin{APACrefauthors}%
Haussler, D.%
\BCBT {}\ \BBA {} Welzl, E.%
\end{APACrefauthors}%
\unskip\
\newblock
\APACrefYearMonthDay{1986}{}{}.
\newblock
{\BBOQ}\APACrefatitle {Epsilon-Nets and Simplex Range Queries} {Epsilon-nets
  and simplex range queries}.{\BBCQ}
\newblock
\BIn{} \APACrefbtitle {Ann. ACM Symp. on Computational Geometry (SoCG).} {Ann.
  acm symp. on computational geometry (socg).}
\PrintBackRefs{\CurrentBib}

\bibitem [\protect \citeauthoryear {%
Hellerstein%
}{%
Hellerstein%
}{%
{\protect \APACyear {2008}}%
}]{%
hellerstein2008parallel}
\APACinsertmetastar {%
hellerstein2008parallel}%
\begin{APACrefauthors}%
Hellerstein, J.%
\end{APACrefauthors}%
\unskip\
\newblock
\APACrefYearMonthDay{2008}{}{}.
\newblock
\APACrefbtitle {Parallel programming in the age of big data. Gigaom Blog. Nov.
  9, 2008.} {Parallel programming in the age of big data. gigaom blog. nov. 9,
  2008.}
\PrintBackRefs{\CurrentBib}

\bibitem [\protect \citeauthoryear {%
Hoeffding%
}{%
Hoeffding%
}{%
{\protect \APACyear {1963}}%
}]{%
hoeffding1963probability}
\APACinsertmetastar {%
hoeffding1963probability}%
\begin{APACrefauthors}%
Hoeffding, W.%
\end{APACrefauthors}%
\unskip\
\newblock
\APACrefYearMonthDay{1963}{}{}.
\newblock
{\BBOQ}\APACrefatitle {Probability inequalities for sums of bounded random
  variables} {Probability inequalities for sums of bounded random
  variables}.{\BBCQ}
\newblock
\APACjournalVolNumPages{Journal of the American statistical
  association}{58}{301}{13--30}.
\PrintBackRefs{\CurrentBib}

\bibitem [\protect \citeauthoryear {%
\APACcitebtitle {IBM: What is big data? Bringing big data to the enterprise}}{%
\APACcitebtitle {IBM: What is big data? Bringing big data to the enterprise}}{%
{\protect \APACyear {2012}}%
}]{%
IBM}
\APACinsertmetastar {%
IBM}%
\APACrefbtitle {IBM: What is big data? Bringing big data to the enterprise.}
  {Ibm: What is big data? bringing big data to the enterprise.}
\newblock
\APACrefYearMonthDay{2012}{}{}.
\newblock
\APAChowpublished {Website}.
\newblock
\APACrefnote{\url{ibm.com/software/data/bigdata/}, accessed on the 3rd of
  October 2012}
\PrintBackRefs{\CurrentBib}

\bibitem [\protect \citeauthoryear {%
Inaba%
, Katoh%
\BCBL {}\ \BBA {} Imai%
}{%
Inaba%
\ \protect \BOthers {.}}{%
{\protect \APACyear {1994}}%
}]{%
IKI94}
\APACinsertmetastar {%
IKI94}%
\begin{APACrefauthors}%
Inaba, M.%
, Katoh, N.%
\BCBL {}\ \BBA {} Imai, H.%
\end{APACrefauthors}%
\unskip\
\newblock
\APACrefYearMonthDay{1994}{}{}.
\newblock
{\BBOQ}\APACrefatitle {Applications of Weighted Voronoi Diagrams and
  Randomization to Variance-Based {\it k}-Clustering} {Applications of weighted
  voronoi diagrams and randomization to variance-based {\it
  k}-clustering}.{\BBCQ}
\newblock
\BIn{} \APACrefbtitle {Symposium on Computational Geometry} {Symposium on
  computational geometry}\ (\BPGS\ 332 -- 339).
\PrintBackRefs{\CurrentBib}

\bibitem [\protect \citeauthoryear {%
Indyk%
, Mahabadi%
, Mahdian%
\BCBL {}\ \BBA {} Mirrokni%
}{%
Indyk%
\ \protect \BOthers {.}}{%
{\protect \APACyear {2014}}%
}]{%
indyk2014composable}
\APACinsertmetastar {%
indyk2014composable}%
\begin{APACrefauthors}%
Indyk, P.%
, Mahabadi, S.%
, Mahdian, M.%
\BCBL {}\ \BBA {} Mirrokni, V\BPBI S.%
\end{APACrefauthors}%
\unskip\
\newblock
\APACrefYearMonthDay{2014}{}{}.
\newblock
{\BBOQ}\APACrefatitle {Composable core-sets for diversity and coverage
  maximization} {Composable core-sets for diversity and coverage
  maximization}.{\BBCQ}
\newblock
\BIn{} \APACrefbtitle {Proceedings of the 33rd ACM SIGMOD-SIGACT-SIGART
  symposium on Principles of database systems} {Proceedings of the 33rd acm
  sigmod-sigact-sigart symposium on principles of database systems}\ (\BPGS\
  100--108).
\PrintBackRefs{\CurrentBib}

\bibitem [\protect \citeauthoryear {%
Joshi%
, Kommaraji%
, Phillips%
\BCBL {}\ \BBA {} Venkatasubramanian%
}{%
Joshi%
\ \protect \BOthers {.}}{%
{\protect \APACyear {2011}}%
}]{%
joshi2011comparing}
\APACinsertmetastar {%
joshi2011comparing}%
\begin{APACrefauthors}%
Joshi, S.%
, Kommaraji, R\BPBI V.%
, Phillips, J\BPBI M.%
\BCBL {}\ \BBA {} Venkatasubramanian, S.%
\end{APACrefauthors}%
\unskip\
\newblock
\APACrefYearMonthDay{2011}{}{}.
\newblock
{\BBOQ}\APACrefatitle {Comparing distributions and shapes using the kernel
  distance} {Comparing distributions and shapes using the kernel
  distance}.{\BBCQ}
\newblock
\BIn{} \APACrefbtitle {Proceedings of the twenty-seventh annual symposium on
  Computational geometry} {Proceedings of the twenty-seventh annual symposium
  on computational geometry}\ (\BPGS\ 47--56).
\PrintBackRefs{\CurrentBib}

\bibitem [\protect \citeauthoryear {%
Langberg%
\ \BBA {} Schulman%
}{%
Langberg%
\ \BBA {} Schulman%
}{%
{\protect \APACyear {2010}}%
}]{%
LS10}
\APACinsertmetastar {%
LS10}%
\begin{APACrefauthors}%
Langberg, M.%
\BCBT {}\ \BBA {} Schulman, L\BPBI J.%
\end{APACrefauthors}%
\unskip\
\newblock
\APACrefYearMonthDay{2010}{}{}.
\newblock
{\BBOQ}\APACrefatitle {Universal epsilon-approximators for Integrals}
  {Universal epsilon-approximators for integrals}.{\BBCQ}
\newblock
\BIn{} \APACrefbtitle {\Proc 21st \SODA} {\proc 21st \soda}\ (\BPGS\ 598--607).
\PrintBackRefs{\CurrentBib}

\bibitem [\protect \citeauthoryear {%
Li%
, Long%
\BCBL {}\ \BBA {} Srinivasan%
}{%
Li%
\ \protect \BOthers {.}}{%
{\protect \APACyear {2001}}%
}]{%
LLS01}
\APACinsertmetastar {%
LLS01}%
\begin{APACrefauthors}%
Li, Y.%
, Long, P\BPBI M.%
\BCBL {}\ \BBA {} Srinivasan, A.%
\end{APACrefauthors}%
\unskip\
\newblock
\APACrefYearMonthDay{2001}{}{}.
\newblock
{\BBOQ}\APACrefatitle {Improved Bounds on the Sample Complexity of Learning}
  {Improved bounds on the sample complexity of learning}.{\BBCQ}
\newblock
\APACjournalVolNumPages{Journal of Computer and System Sciences
  (JCSS)}{62}{}{516--527}.
\PrintBackRefs{\CurrentBib}

\bibitem [\protect \citeauthoryear {%
Liberty%
}{%
Liberty%
}{%
{\protect \APACyear {2017}}%
}]{%
code1}
\APACinsertmetastar {%
code1}%
\begin{APACrefauthors}%
Liberty, E.%
\end{APACrefauthors}%
\unskip\
\newblock
\APACrefYearMonthDay{2017}{}{}.
\newblock
\APACrefbtitle {Sketches for Frequent Directions,
  {https://edoliberty.github.io/}} {Sketches for frequent directions,
  {https://edoliberty.github.io/}}\ \APACbVolEdTR{}{\BTR{}}.
\newblock
\APACaddressInstitution{}{Amazon LTD}.
\PrintBackRefs{\CurrentBib}

\bibitem [\protect \citeauthoryear {%
L{\"o}ffler%
\ \BBA {} Phillips%
}{%
L{\"o}ffler%
\ \BBA {} Phillips%
}{%
{\protect \APACyear {2009}}%
}]{%
loffler2009shape}
\APACinsertmetastar {%
loffler2009shape}%
\begin{APACrefauthors}%
L{\"o}ffler, M.%
\BCBT {}\ \BBA {} Phillips, J\BPBI M.%
\end{APACrefauthors}%
\unskip\
\newblock
\APACrefYearMonthDay{2009}{}{}.
\newblock
{\BBOQ}\APACrefatitle {Shape fitting on point sets with probability
  distributions} {Shape fitting on point sets with probability
  distributions}.{\BBCQ}
\newblock
\BIn{} \APACrefbtitle {European Symposium on Algorithms} {European symposium on
  algorithms}\ (\BPGS\ 313--324).
\PrintBackRefs{\CurrentBib}

\bibitem [\protect \citeauthoryear {%
Maalouf%
, Jubran%
\BCBL {}\ \BBA {} Feldman%
}{%
Maalouf%
\ \protect \BOthers {.}}{%
{\protect \APACyear {2019}}%
}]{%
alaa}
\APACinsertmetastar {%
alaa}%
\begin{APACrefauthors}%
Maalouf, A.%
, Jubran, I.%
\BCBL {}\ \BBA {} Feldman, D.%
\end{APACrefauthors}%
\unskip\
\newblock
\APACrefYearMonthDay{2019}{}{}.
\newblock
{\BBOQ}\APACrefatitle {Fast and Accurate Least-Mean-Squares Solvers} {Fast and
  accurate least-mean-squares solvers}.{\BBCQ}
\newblock
\APACjournalVolNumPages{arXiv preprint 1906.04705}{}{}{}.
\PrintBackRefs{\CurrentBib}

\bibitem [\protect \citeauthoryear {%
Mahajan%
, Nimbhorkar%
\BCBL {}\ \BBA {} Varadarajan%
}{%
Mahajan%
\ \protect \BOthers {.}}{%
{\protect \APACyear {2009}}%
}]{%
mahajan2009planar}
\APACinsertmetastar {%
mahajan2009planar}%
\begin{APACrefauthors}%
Mahajan, M.%
, Nimbhorkar, P.%
\BCBL {}\ \BBA {} Varadarajan, K.%
\end{APACrefauthors}%
\unskip\
\newblock
\APACrefYearMonthDay{2009}{}{}.
\newblock
{\BBOQ}\APACrefatitle {The planar k-means problem is NP-hard} {The planar
  k-means problem is np-hard}.{\BBCQ}
\newblock
\BIn{} \APACrefbtitle {International Workshop on Algorithms and Computation}
  {International workshop on algorithms and computation}\ (\BPGS\ 274--285).
\PrintBackRefs{\CurrentBib}

\bibitem [\protect \citeauthoryear {%
Mahoney%
}{%
Mahoney%
}{%
{\protect \APACyear {2011}}%
}]{%
Mahoney}
\APACinsertmetastar {%
Mahoney}%
\begin{APACrefauthors}%
Mahoney, M\BPBI W.%
\end{APACrefauthors}%
\unskip\
\newblock
\APACrefYearMonthDay{2011}{}{}.
\newblock
{\BBOQ}\APACrefatitle {Randomized Algorithms for Matrices and Data} {Randomized
  algorithms for matrices and data}.{\BBCQ}
\newblock
\APACjournalVolNumPages{Foundations and Trends in Machine
  Learning}{3}{2}{123--224}.
\PrintBackRefs{\CurrentBib}

\bibitem [\protect \citeauthoryear {%
Matouaek%
}{%
Matouaek%
}{%
{\protect \APACyear {2003}}%
}]{%
matouaek2003new}
\APACinsertmetastar {%
matouaek2003new}%
\begin{APACrefauthors}%
Matouaek, J.%
\end{APACrefauthors}%
\unskip\
\newblock
\APACrefYearMonthDay{2003}{}{}.
\newblock
{\BBOQ}\APACrefatitle {New constructions of weak epsilon-nets} {New
  constructions of weak epsilon-nets}.{\BBCQ}
\newblock
\BIn{} \APACrefbtitle {Proceedings of the nineteenth annual symposium on
  Computational geometry} {Proceedings of the nineteenth annual symposium on
  computational geometry}\ (\BPGS\ 129--135).
\PrintBackRefs{\CurrentBib}

\bibitem [\protect \citeauthoryear {%
Matousek%
}{%
Matousek%
}{%
{\protect \APACyear {1995}}%
}]{%
Matousek95}
\APACinsertmetastar {%
Matousek95}%
\begin{APACrefauthors}%
Matousek, J.%
\end{APACrefauthors}%
\unskip\
\newblock
\APACrefYearMonthDay{1995}{}{}.
\newblock
{\BBOQ}\APACrefatitle {Approximations and Optimal Geometric Divide-an-Conquer}
  {Approximations and optimal geometric divide-an-conquer}.{\BBCQ}
\newblock
\APACjournalVolNumPages{J. Comput. Syst. Sci}{50}{2}{203--208}.
\PrintBackRefs{\CurrentBib}

\bibitem [\protect \citeauthoryear {%
McLachlan%
\ \BBA {} Krishnan%
}{%
McLachlan%
\ \BBA {} Krishnan%
}{%
{\protect \APACyear {2007}}%
}]{%
mclachlan2007algorithm}
\APACinsertmetastar {%
mclachlan2007algorithm}%
\begin{APACrefauthors}%
McLachlan, G.%
\BCBT {}\ \BBA {} Krishnan, T.%
\end{APACrefauthors}%
\unskip\
\newblock
\APACrefYear{2007}.
\newblock
\APACrefbtitle {The EM algorithm and extensions} {The em algorithm and
  extensions}\ (\BVOL~382).
\newblock
\APACaddressPublisher{}{John Wiley \& Sons}.
\PrintBackRefs{\CurrentBib}

\bibitem [\protect \citeauthoryear {%
Munteanu%
\ \BBA {} Schwiegelshohn%
}{%
Munteanu%
\ \BBA {} Schwiegelshohn%
}{%
{\protect \APACyear {2018}}%
}]{%
munteanu2018coresets}
\APACinsertmetastar {%
munteanu2018coresets}%
\begin{APACrefauthors}%
Munteanu, A.%
\BCBT {}\ \BBA {} Schwiegelshohn, C.%
\end{APACrefauthors}%
\unskip\
\newblock
\APACrefYearMonthDay{2018}{}{}.
\newblock
{\BBOQ}\APACrefatitle {Coresets-Methods and History: A Theoreticians Design
  Pattern for Approximation and Streaming Algorithms} {Coresets-methods and
  history: A theoreticians design pattern for approximation and streaming
  algorithms}.{\BBCQ}
\newblock
\APACjournalVolNumPages{KI-K{\"u}nstliche Intelligenz}{32}{1}{37--53}.
\PrintBackRefs{\CurrentBib}

\bibitem [\protect \citeauthoryear {%
Muthukrishnan%
\ \protect \BOthers {.}}{%
Muthukrishnan%
\ \protect \BOthers {.}}{%
{\protect \APACyear {2005}}%
}]{%
muthukrishnan2005data}
\APACinsertmetastar {%
muthukrishnan2005data}%
\begin{APACrefauthors}%
Muthukrishnan, S.%
\BCBT {}\ \BOthersPeriod {.}
\end{APACrefauthors}%
\unskip\
\newblock
\APACrefYearMonthDay{2005}{}{}.
\newblock
{\BBOQ}\APACrefatitle {Data streams: Algorithms and applications} {Data
  streams: Algorithms and applications}.{\BBCQ}
\newblock
\APACjournalVolNumPages{Foundations and Trends{\textregistered} in Theoretical
  Computer Science}{1}{2}{117--236}.
\PrintBackRefs{\CurrentBib}

\bibitem [\protect \citeauthoryear {%
Nasser%
, Jubran%
\BCBL {}\ \BBA {} Feldman%
}{%
Nasser%
\ \protect \BOthers {.}}{%
{\protect \APACyear {2015}}%
}]{%
nasser2015low}
\APACinsertmetastar {%
nasser2015low}%
\begin{APACrefauthors}%
Nasser, S.%
, Jubran, I.%
\BCBL {}\ \BBA {} Feldman, D.%
\end{APACrefauthors}%
\unskip\
\newblock
\APACrefYearMonthDay{2015}{}{}.
\newblock
{\BBOQ}\APACrefatitle {Low-cost and Faster Tracking Systems Using Core-sets for
  Pose-Estimation} {Low-cost and faster tracking systems using core-sets for
  pose-estimation}.{\BBCQ}
\newblock
\APACjournalVolNumPages{arXiv preprint arXiv:1511.09120}{}{}{}.
\PrintBackRefs{\CurrentBib}

\bibitem [\protect \citeauthoryear {%
Ostrovsky%
, Rabani%
, Schulman%
\BCBL {}\ \BBA {} Swamy%
}{%
Ostrovsky%
\ \protect \BOthers {.}}{%
{\protect \APACyear {2006}}%
}]{%
ostrovsky2006effectiveness}
\APACinsertmetastar {%
ostrovsky2006effectiveness}%
\begin{APACrefauthors}%
Ostrovsky, R.%
, Rabani, Y.%
, Schulman, L\BPBI J.%
\BCBL {}\ \BBA {} Swamy, C.%
\end{APACrefauthors}%
\unskip\
\newblock
\APACrefYearMonthDay{2006}{}{}.
\newblock
{\BBOQ}\APACrefatitle {The effectiveness of Lloyd-type methods for the k-means
  problem} {The effectiveness of lloyd-type methods for the k-means
  problem}.{\BBCQ}
\newblock
\BIn{} \APACrefbtitle {Foundations of Computer Science, 2006. FOCS'06. 47th
  Annual IEEE Symposium on} {Foundations of computer science, 2006. focs'06.
  47th annual ieee symposium on}\ (\BPGS\ 165--176).
\PrintBackRefs{\CurrentBib}

\bibitem [\protect \citeauthoryear {%
Paul%
, Feldman%
, Rus%
\BCBL {}\ \BBA {} Newman%
}{%
Paul%
\ \protect \BOthers {.}}{%
{\protect \APACyear {2014}}%
}]{%
paul2014visual}
\APACinsertmetastar {%
paul2014visual}%
\begin{APACrefauthors}%
Paul, R.%
, Feldman, D.%
, Rus, D.%
\BCBL {}\ \BBA {} Newman, P.%
\end{APACrefauthors}%
\unskip\
\newblock
\APACrefYearMonthDay{2014}{}{}.
\newblock
{\BBOQ}\APACrefatitle {Visual precis generation using coresets} {Visual precis
  generation using coresets}.{\BBCQ}
\newblock
\BIn{} \APACrefbtitle {Robotics and Automation (ICRA), 2014 IEEE International
  Conference on} {Robotics and automation (icra), 2014 ieee international
  conference on}\ (\BPGS\ 1304--1311).
\PrintBackRefs{\CurrentBib}

\bibitem [\protect \citeauthoryear {%
Phillips%
}{%
Phillips%
}{%
{\protect \APACyear {2016}}%
}]{%
Phillips16}
\APACinsertmetastar {%
Phillips16}%
\begin{APACrefauthors}%
Phillips, J\BPBI M.%
\end{APACrefauthors}%
\unskip\
\newblock
\APACrefYearMonthDay{2016}{}{}.
\newblock
{\BBOQ}\APACrefatitle {Coresets and Sketches, Near-final version of Chapter 49
  in Handbook on Discrete and Computational Geometry, 3rd edition} {Coresets
  and sketches, near-final version of chapter 49 in handbook on discrete and
  computational geometry, 3rd edition}.{\BBCQ}
\newblock
\APACjournalVolNumPages{CoRR}{abs/1601.00617}{}{}.
\newblock
\begin{APACrefURL} \url{http://arxiv.org/abs/1601.00617} \end{APACrefURL}
\PrintBackRefs{\CurrentBib}

\bibitem [\protect \citeauthoryear {%
Rosman%
, Volkov%
, Feldman%
, Fisher~III%
\BCBL {}\ \BBA {} Rus%
}{%
Rosman%
\ \protect \BOthers {.}}{%
{\protect \APACyear {2014}}%
}]{%
rosman2014coresets}
\APACinsertmetastar {%
rosman2014coresets}%
\begin{APACrefauthors}%
Rosman, G.%
, Volkov, M.%
, Feldman, D.%
, Fisher~III, J\BPBI W.%
\BCBL {}\ \BBA {} Rus, D.%
\end{APACrefauthors}%
\unskip\
\newblock
\APACrefYearMonthDay{2014}{}{}.
\newblock
{\BBOQ}\APACrefatitle {Coresets for k-segmentation of streaming data} {Coresets
  for k-segmentation of streaming data}.{\BBCQ}
\newblock
\BIn{} \APACrefbtitle {Advances in Neural Information Processing Systems
  (NIPS)} {Advances in neural information processing systems (nips)}\ (\BPGS\
  559--567).
\PrintBackRefs{\CurrentBib}

\bibitem [\protect \citeauthoryear {%
Sanabria%
}{%
Sanabria%
}{%
{\protect \APACyear {2018}}%
}]{%
code4}
\APACinsertmetastar {%
code4}%
\begin{APACrefauthors}%
Sanabria, J\BPBI M.%
\end{APACrefauthors}%
\unskip\
\newblock
\APACrefYearMonthDay{2018}{}{}.
\newblock
\APACrefbtitle {Randomized matrix algorithms
  {https://github.com/jomsdev/randNLA}} {Randomized matrix algorithms
  {https://github.com/jomsdev/randNLA}}\ \APACbVolEdTR{}{\BTR{}}.
\newblock
\APACaddressInstitution{}{TravelPerk}.
\PrintBackRefs{\CurrentBib}

\bibitem [\protect \citeauthoryear {%
Segaran%
\ \BBA {} Hammerbacher%
}{%
Segaran%
\ \BBA {} Hammerbacher%
}{%
{\protect \APACyear {2009}}%
}]{%
segaran2009beautiful}
\APACinsertmetastar {%
segaran2009beautiful}%
\begin{APACrefauthors}%
Segaran, T.%
\BCBT {}\ \BBA {} Hammerbacher, J.%
\end{APACrefauthors}%
\unskip\
\newblock
\APACrefYear{2009}.
\newblock
\APACrefbtitle {Beautiful data: the stories behind elegant data solutions}
  {Beautiful data: the stories behind elegant data solutions}.
\newblock
\APACaddressPublisher{}{" O'Reilly Media, Inc."}.
\PrintBackRefs{\CurrentBib}

\bibitem [\protect \citeauthoryear {%
Sener%
\ \BBA {} Savarese%
}{%
Sener%
\ \BBA {} Savarese%
}{%
{\protect \APACyear {2017}}%
}]{%
sener2017active}
\APACinsertmetastar {%
sener2017active}%
\begin{APACrefauthors}%
Sener, O.%
\BCBT {}\ \BBA {} Savarese, S.%
\end{APACrefauthors}%
\unskip\
\newblock
\APACrefYearMonthDay{2017}{}{}.
\newblock
{\BBOQ}\APACrefatitle {Active learning for CONVOLUTIONAL NEURAL NETWORKS: A
  CORE-SET APPROACH} {Active learning for convolutional neural networks: A
  core-set approach}.{\BBCQ}
\newblock
\APACjournalVolNumPages{stat}{1050}{}{27}.
\PrintBackRefs{\CurrentBib}

\bibitem [\protect \citeauthoryear {%
Shyamalkumar%
\ \BBA {} Varadarajan%
}{%
Shyamalkumar%
\ \BBA {} Varadarajan%
}{%
{\protect \APACyear {2007}}%
}]{%
SV07}
\APACinsertmetastar {%
SV07}%
\begin{APACrefauthors}%
Shyamalkumar, N.%
\BCBT {}\ \BBA {} Varadarajan, K.%
\end{APACrefauthors}%
\unskip\
\newblock
\APACrefYearMonthDay{2007}{}{}.
\newblock
{\BBOQ}\APACrefatitle {Efficient subspace approximation algorithms} {Efficient
  subspace approximation algorithms}.{\BBCQ}
\newblock
\BIn{} \APACrefbtitle {\Proc 18th \SODA} {\proc 18th \soda}\ (\BPGS\ 532--540).
\PrintBackRefs{\CurrentBib}

\bibitem [\protect \citeauthoryear {%
Tolochinsky%
\ \BBA {} Feldman%
}{%
Tolochinsky%
\ \BBA {} Feldman%
}{%
{\protect \APACyear {2018}}%
}]{%
tolochinsky2018coresets}
\APACinsertmetastar {%
tolochinsky2018coresets}%
\begin{APACrefauthors}%
Tolochinsky, E.%
\BCBT {}\ \BBA {} Feldman, D.%
\end{APACrefauthors}%
\unskip\
\newblock
\APACrefYearMonthDay{2018}{}{}.
\newblock
{\BBOQ}\APACrefatitle {Coresets For Monotonic Functions with Applications to
  Deep Learning} {Coresets for monotonic functions with applications to deep
  learning}.{\BBCQ}
\newblock
\APACjournalVolNumPages{arXiv preprint arXiv:1802.07382}{}{}{}.
\PrintBackRefs{\CurrentBib}

\bibitem [\protect \citeauthoryear {%
Tutz%
\ \BBA {} Binder%
}{%
Tutz%
\ \BBA {} Binder%
}{%
{\protect \APACyear {2007}}%
}]{%
tutz2007boosting}
\APACinsertmetastar {%
tutz2007boosting}%
\begin{APACrefauthors}%
Tutz, G.%
\BCBT {}\ \BBA {} Binder, H.%
\end{APACrefauthors}%
\unskip\
\newblock
\APACrefYearMonthDay{2007}{}{}.
\newblock
{\BBOQ}\APACrefatitle {Boosting ridge regression} {Boosting ridge
  regression}.{\BBCQ}
\newblock
\APACjournalVolNumPages{Computational Statistics \& Data
  Analysis}{51}{12}{6044--6059}.
\PrintBackRefs{\CurrentBib}

\bibitem [\protect \citeauthoryear {%
Varadarajan%
\ \BBA {} Xiao%
}{%
Varadarajan%
\ \BBA {} Xiao%
}{%
{\protect \APACyear {2012}}%
{\protect \APACexlab {{\protect \BCnt {1}}}}}]{%
VX12-soda}
\APACinsertmetastar {%
VX12-soda}%
\begin{APACrefauthors}%
Varadarajan, K.%
\BCBT {}\ \BBA {} Xiao, X.%
\end{APACrefauthors}%
\unskip\
\newblock
\APACrefYearMonthDay{2012{\protect \BCnt {1}}}{}{}.
\newblock
{\BBOQ}\APACrefatitle {A near-linear algorithm for projective clustering
  integer points} {A near-linear algorithm for projective clustering integer
  points}.{\BBCQ}
\newblock
\BIn{} \APACrefbtitle {\Proc \SODA.} {\proc \soda.}
\PrintBackRefs{\CurrentBib}

\bibitem [\protect \citeauthoryear {%
Varadarajan%
\ \BBA {} Xiao%
}{%
Varadarajan%
\ \BBA {} Xiao%
}{%
{\protect \APACyear {2012}}%
{\protect \APACexlab {{\protect \BCnt {2}}}}}]{%
varadarajan}
\APACinsertmetastar {%
varadarajan}%
\begin{APACrefauthors}%
Varadarajan, K.%
\BCBT {}\ \BBA {} Xiao, X.%
\end{APACrefauthors}%
\unskip\
\newblock
\APACrefYearMonthDay{2012{\protect \BCnt {2}}}{}{}.
\newblock
{\BBOQ}\APACrefatitle {On the sensitivity of shape fitting problems} {On the
  sensitivity of shape fitting problems}.{\BBCQ}
\newblock
\BIn{} \APACrefbtitle {\Proc 32nd Annual Conference on \FSTTCS} {\proc 32nd
  annual conference on \fsttcs}\ (\BPGS\ 486 -- 497).
\PrintBackRefs{\CurrentBib}

\bibitem [\protect \citeauthoryear {%
Yahoo!%
}{%
Yahoo!%
}{%
{\protect \APACyear {2018}}%
}]{%
code5}
\APACinsertmetastar {%
code5}%
\begin{APACrefauthors}%
Yahoo!%
\end{APACrefauthors}%
\unskip\
\newblock
\APACrefYearMonthDay{2018}{}{}.
\newblock
\APACrefbtitle {Data sketches, {https://datasketches.github.io/}} {Data
  sketches, {https://datasketches.github.io/}}\ \APACbVolEdTR{}{\BTR{}}.
\newblock
\APACaddressInstitutionEqAuth{}{Yahoo!}
\PrintBackRefs{\CurrentBib}

\end{thebibliography}

\end{document}